%% file: main.tex
\documentclass[10pt,twocolumn,letterpaper]{article}

\usepackage[pagenumbers]{iccv} 


\definecolor{iccvblue}{rgb}{0.21,0.49,0.74}
\usepackage[pagebackref,breaklinks,colorlinks,allcolors=iccvblue]{hyperref}
\usepackage{multirow}
\usepackage{pifont}
\usepackage{algorithm}
\usepackage{listings}
\usepackage{algpseudocode}
\def\paperID{14407} 
\def\confName{ICCV}
\def\confYear{2025}

\title{HybridGen: VLM-Guided Hybrid Planning for Scalable Data Generation of Imitation Learning 
}

\author{
Wensheng Wang \qquad Ning Tan \\
Sun Yat-sen University, Guangzhou, China \\
{\tt\small wangwsh23@mail2.sysu.edu.cn, tann5@mail.sysu.edu.cn}
}
\begin{document}
\maketitle
\input{sec/0_abstract}    
\input{sec/1_intro}
\input{sec/2_relatedwork}
\input{sec/3_approach}
\input{sec/4_experiments}

\input{sec/5_conclusion}
{
    \small
    \bibliographystyle{ieeenat_fullname}
    \bibliography{main}
}
\input{sec/X_suppl}

\end{document}

%% file: sec/0_abstract.tex
\begin{abstract}
The acquisition of large-scale and diverse demonstration data are essential for improving robotic imitation learning generalization. However, generating such data for complex manipulations is challenging in real-world settings. We introduce HybridGen, an automated framework that integrates Vision-Language Model (VLM) and hybrid planning. HybridGen uses a two-stage pipeline: first, VLM to parse expert demonstrations, decomposing tasks into expert-dependent (object-centric pose transformations for precise control) and plannable segments (synthesizing diverse trajectories via path planning); second, pose transformations substantially expand the first-stage data. Crucially, HybridGen generates a large volume of training data without requiring specific data formats, making it broadly applicable to a wide range of imitation learning algorithms, a characteristic which we also demonstrate empirically across multiple algorithms. Evaluations across seven tasks and their variants demonstrate that agents trained with HybridGen achieve substantial performance and generalization gains, averaging a 5\% improvement over state-of-the-art methods. Notably, in the most challenging task variants, HybridGen achieves significant improvement, reaching a 59.7\% average success rate, significantly outperforming Mimicgen's 49.5\%. These results demonstrating its effectiveness and practicality.
\end{abstract}

%% file: sec/1_intro.tex
\section{Introduction}
\begin{figure*}
    \centering
     \includegraphics[width=1\linewidth]{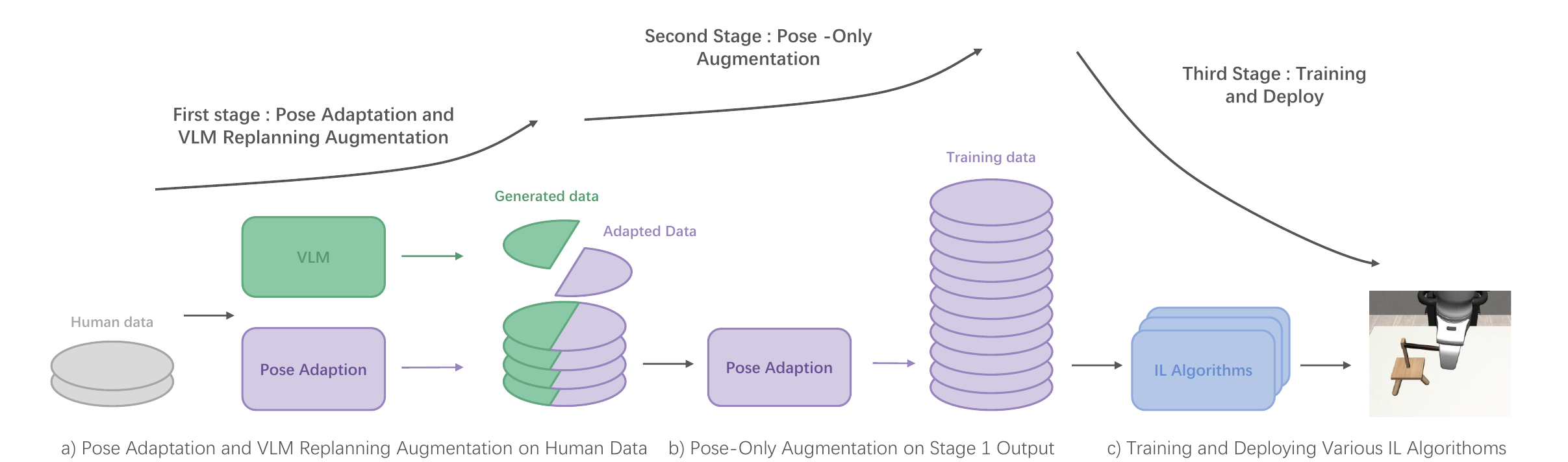}
     \caption{\textbf{HybridGen Overview}. We propose an automated data generation framework leveraging VLM and hybrid planning strategies. (a) The first stage involves the initial augmentation of limited human data using pose adaptation and VLM replanning. (b) The second stage performs further augmentation through pose-only adaptation, resulting in a larger training dataset. (c) The data generated by this framework can be seamlessly used to train various imitation learning algorithms.}
     \label{fig:onecol}
  \end{figure*}
The key challenge in robot imitation learning lies in acquiring broad and diverse demonstration data with efficient collection costs. While large-scale human demonstration datasets have proven effective in enhancing policy generalization capabilities\cite{kimContextAwarePlanningEnvironmentAware2024, linDATASCALINGLAWS, zhaoALOHAUnleashedSimple, zhangLeveragingLocalityBoost2024}, their acquisition often incurs substantial time and economic overhead. Complex robotic tasks, such as intricate assembly or dexterous manipulation of deformable objects, often require a nuanced understanding of the environment and the task goals, making data collection labor-intensive and time-consuming. Meanwhile, internet-scale multimodal data endows VLMs with powerful scene understanding abilities\cite{openaiGPT4TechnicalReport2024a, teamGemini15Unlocking2024}. Yet direct application of VLMs to robotic trajectory generation faces a fundamental bottleneck: VLMs lack explicit modeling of robot kinematics and dynamic constraints, limiting their output to coarse task descriptions rather than providing trajectory-level guidance with execution-level precision. Existing data augmentation methods, though capable of spatial expansion through pose transformations or motion planning\cite{mandlekarMimicGenDataGeneration2023, garrettSkillMimicGenAutomatedDemonstration2024, mandlekarHumanintheLoopTaskMotion2023}, struggle to generate substantially diverse demonstrations at the task semantics level.

To address these challenges, we introduce \textbf{HybridGen}, an innovative automated framework designed to facilitate the scalable generation of imitation learning data. Our approach strategically integrates the power of VLM with a hybrid planning methodology.  HybridGen operates through a two-phase process: first, it leverages VLM to dissect expert demonstrations, effectively decoupling tasks into segments that demand precise, expert-guided control (object-centric pose transformations) and those amenable to automated planning (generation of diverse trajectories).  Subsequently, pose transformations are applied to substantially augment the initial dataset.

The strength of HybridGen lies in its ability to generate format-independent, high-quality training data. This format independence ensures broad applicability across diverse imitation learning algorithms, a key advantage we empirically validate across multiple algorithms. Our extensive evaluations across seven diverse robotic manipulation tasks and their difficulty variants demonstrate HybridGen's superior performance.  Agents trained on HybridGen-generated data achieve an average 5\% improvement in task success rates compared to state-of-the-art methods.  Notably, even in challenging task variants, HybridGen achieves a 59.7\% average success rate, significantly outperforming Mimicgen's 49.5\%.  These agents exhibit not only performance gains but also enhanced generalization.  Furthermore, we validated the broad applicability of HybridGen-generated data across various imitation learning algorithms, including Behavioral Cloning(BC)\cite{pomerleau1988alvinn} with an RNN policy and BC with a Transformer policy(all from Robomimic\cite{mandlekar2022matters}), as well as Diffusion Policy\cite{chiDiffusionPolicyVisuomotor2024}, demonstrating its algorithmic robustness.
In summary, our main contributions are the following:
\begin{itemize}
    \item We present \textbf{HybridGen}, a novel automated framework for scalable imitation learning data generation, which uniquely integrates VLM with a hybrid planning approach.
    \item HybridGen generates data in a \textbf{format consistent with human demonstrations}, ensuring broad applicability across various imitation learning algorithms and facilitating seamless integration with existing pipelines.
    \item Our evaluations demonstrate \textbf{state-of-the-art success rates} (59.7\% vs. 49.5\% baseline) on the most challenging task variants, and with a consistent 5\% average improvement across all tasks, showcasing superior generalization to unseen scenarios.
\end{itemize}

%% file: sec/2_relatedwork.tex
\section{Related Work}
\label{sec:related}
\textbf{VLMs and LLMs for robotics.}
Vision-Language Models (VLMs) and Large Language Models (LLMs) have demonstrated strong capabilities in high-level task planning and generalization. Some works \cite{huLookYouLeap2023, huangInstruct2ActMappingMultimodality2023, jiangVIMAGeneralRobot2023, liangCodePoliciesLanguage2023, wangVLMSeeRobot2024, wuTidyBotPersonalizedRobot} leverage the rich knowledge learned by VLMs and LLMs from vast datasets. These models enable high-level robot control through requirement understanding, task planning, and action generation. However, directly translating language to low-level robot actions remains a significant challenge. Several recent works have explored using VLMs to guide motion planning. For instance, VoxPoser \cite{huangVoxPoserComposable3D2023} uses a VLM to extract spatial relationships and affordances from a scene, which are then used to formulate a constrained optimization problem to guide low-level robot control in open-world scenarios. Similarly, some works \cite{huangCoPaGeneralRobotic, huangReKepSpatioTemporalReasoning, tangKALIEFineTuningVisionLanguage2024, zhouCodeasMonitorConstraintawareVisual2024} use VLMs to identify keypoints or regions of interest in images, guiding the robot's manipulation. While these methods can achieve basic task completion, their limited spatial reasoning capabilities hinder their effectiveness in complex, precision-demanding tasks.We address this issue by using adapted expert demonstrations in high-precision demanding phases.
\\\\\textbf{VLMs and LLMs for data generation.}
Another line of research employs VLMs and LLMs for simulation-based data generation. RoboGen \cite{wangRoboGenUnleashingInfinite2023} and RobotWin \cite{muRoboTwinDualArmRobot2024} use 3D generative models and LLMs to create diverse data. Scaling Up and Distilling Down \cite{haScalingDistillingLanguageGuided2023} and RobotGPT \cite{jinRobotGPTRobotManipulation2023} use LLMs to generate manipulation policies, execute them, and collect trajectories for agent training. However many of these approaches rely heavily on predefined motion primitives, thereby failing to address intricate manipulation requirements. Our framework addresses these limitations by strategically combining VLM-based planning for non-critical phases with expert demonstrations for mission-critical stages, simultaneously leveraging VLMs to enhance data diversity while collecting complex operational data.
\\\\\textbf{Coarse-to-Fine Task Decomposition.}
Existing approaches in robot manipulation often decompose tasks into coarse and fine-grained phases to leverage both automated planning and human expertise. In typical frameworks \cite{garrettSkillMimicGenAutomatedDemonstration2024, paloLearningMultiStageTasks2021,johns2021coarse,vosylius2023start}, the coarse phase (e.g., object alignment) is addressed via motion planning, while the fine-grained phase (e.g., precise interactions) relies on expert demonstrations. This paradigm efficiently utilizes human data while partially automating data collection, significantly reducing human effort \cite{mandlekarHumanintheLoopTaskMotion2023}. However, existing methods often adopt simplistic motion planners (e.g., interpolation or inverse kinematics), limiting trajectory diversity and adaptability to unseen scenarios. In contrast to methods that utilize basic interpolation or inverse kinematics for motion planning, our approach employs a Vision-Language Model to extract relevant constraints. These constraints then guide trajectory generation, leading to more reasonable and diverse robot motions.
\\\\\textbf{Data Augmentation for Imitation Learning.}
Conventional data augmentation approaches \cite{garrettSkillMimicGenAutomatedDemonstration2024, jiangDexMimicGenAutomatedData2024, mandlekarMimicGenDataGeneration2023} primarily apply human demonstrations to new scenes through pose variations, offering limited diversity through mere environmental replay or simple trajectory interpolation. In contrast, our method fully exploits VLM-derived knowledge by regenerating partial trajectories under constraints proposed by VLMs. This achieves substantial diversity enhancement through semantically-informed trajectory restructuring. Critically, our framework mitigates collisions and avoids unsuitable initial poses by strategically selecting optimal sub-segments of the generated trajectories.

%% file: sec/3_approach.tex
\section{Approach}
In this section, we introduce how Hybridgen performs two-stage data augmentation on a small number of human demonstrations to obtain large-scale high-quality demonstrations. In  Sec. \ref{task_decomposition}, we explain how to decompose each demonstration trajectory at two granularities: 1) subtask-level decomposition and 2) pose-level decomposition.  Sec. \ref{first} introduces the first augmentation stage that combines pose transformation with VLM-proposed constraints to expand human demonstrations. Sec. \ref{second} presents the second augmentation stage that further scales up the initially augmented data. Sec. \ref{constraint} details our method for generating constraints and planning paths using VLMs.
\subsection{Task Decomposition}

\label{task_decomposition}
For each task, we prepare a small number of successful source demonstrations operated by humans, and segment each task into several object-centric subtasks as in MimicGen. Each demonstrated task is assumed to be composed of a sequence of distinct, object-centric subtasks. To analyze these demonstrations, we process each trajectory $\tau$ from our source dataset and divide it into $M$ subtask segments $\{\tau_i\}_{i=1}^{M}$. Each segment, denoted as $\tau_i$, is designed to correspond to a specific subtask $S_i(o_{s_i})$, thus decomposing the trajectory $\tau$ into an ordered set of segments $\tau = (\tau_1, \tau_2, \ldots, \tau_M)$, where each segment represents a single, coherent subtask. Furthermore, to achieve finer control over trajectory generation, we categorize each pose $T$ within the entire trajectory $\tau$. Regardless of which subtask segment $\tau_i$ a pose $T$ belongs to, we classify it as either a data-dependent pose ($T^D$) or a replanning pose ($T^R$). Data-dependent poses $T^D$ appear in complex trajectories, require high precision, and are directly derived from the source data. Examples include poses in intricate manipulations like inserting a building block into another. Replanning poses $T^R$ appear in simpler trajectories, such as translations, rotations, or grasping actions, which can be effectively generated by our trajectory planner. Therefore, the trajectory $\tau$ can be represented as a sequence that encompasses both subtask segment structure and pose type information: Firstly, the trajectory is segmented into $M$ subtask segments $\tau = (\tau_1, \tau_2, \ldots, \tau_M)$; secondly, within the entire trajectory $\tau$, each pose is labeled with a type, ultimately represented as $\tau = (T_1^{type_1}, T_2^{type_2}, \ldots, T_N^{type_N})$, where $type_j \in \{D, R\}$ indicates whether the $j$-th pose is data-dependent or a replanning pose. We use Gemini to analyze the video recorded for each source demonstration and output the coordinates of the source data-dependent poses in each video in a structured frame-by-frame format. To address the limited precision of Gemini, we interpolate each frame of the video by 10 times before analysis and map the obtained dependent pose coordinates to the subscript of the corresponding pose $T_i$.

\subsection{First augmentation}
\label{first}
\begin{figure*}
    \centering
     \includegraphics[width=1\linewidth]{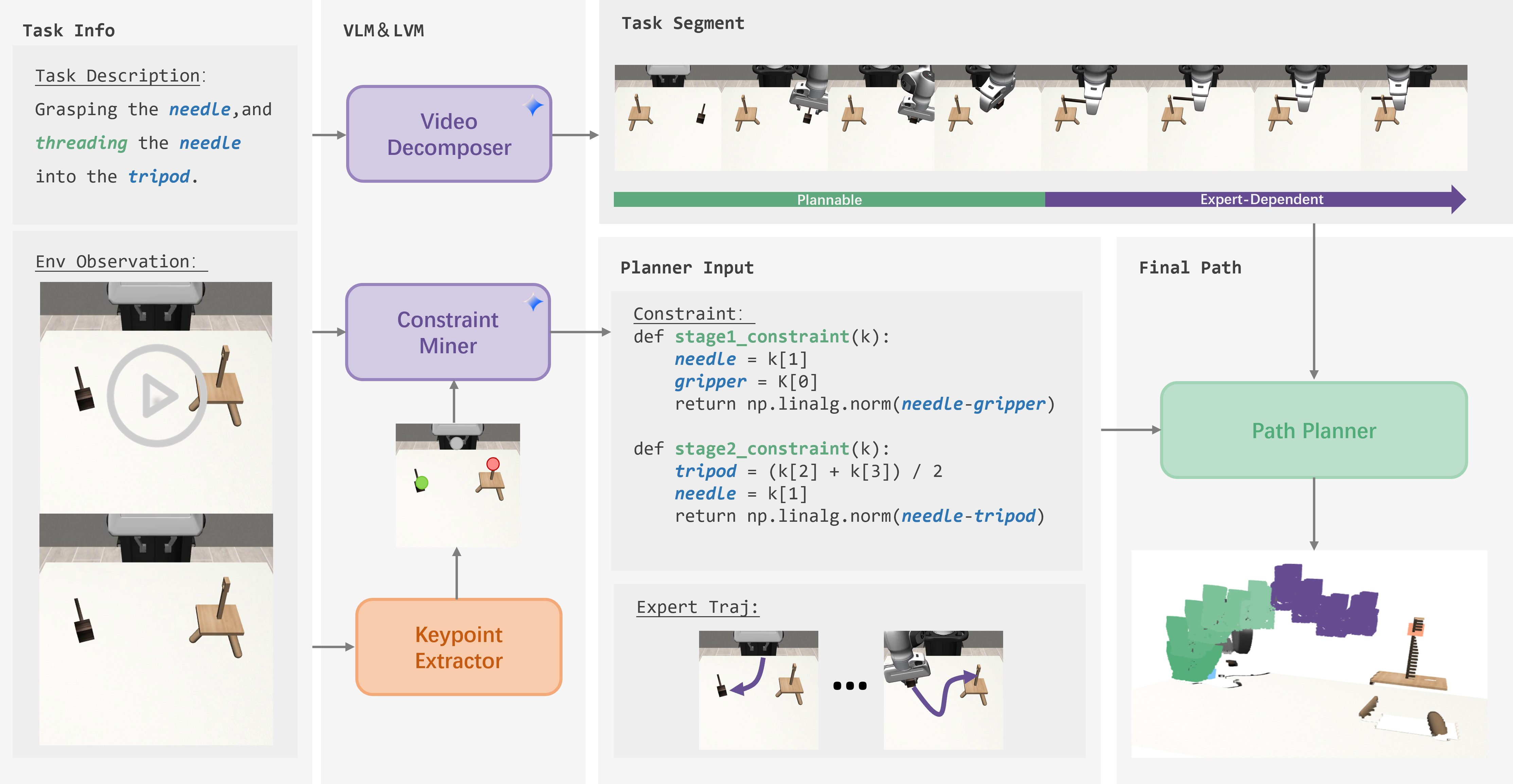}
     \caption{\textbf{Pipeline}. For each task, HybridGen takes the following inputs: a textual task description, a video recording of a human demonstration, and an initial RGB image of the environment. The Video Decomposer parses the task , dividing the complete trajectory into Plannable and Expert-Dependent segments. The Keypoint Extractor extract task-relevant keypoints. The Constraint Miner module takes the environment's RGB image (annotated with keypoints) and the task description, and generates the constraints required to accomplish the task. The Path Planner first calculates the transformed expert demonstrations for the current environment based on the Expert demonstrations data, then replans the Plannable segments according to the derived constraints. Through this process, HybridGen integrates expert demonstrations information with the prior knowledge of a VLM, significantly enhancing the diversity of the generated demonstrations.}
     \label{fig:pipeline}
  \end{figure*}
In the first data augmentation step, we fully utilize a small source dataset, combined with a VLM, to generate a substantial and trajectory-rich dataset $D_{src}^{'}$ for subsequent augmentation.For a task composed of M subtasks, we select subtask trajectories from the source dataset using a suitable selection strategy (detailed in Sec. \ref{second}), and apply pose adaptation and replanning to these trajectories.
\\\\\textbf{Data-dependent segments.}
For data-dependent segments denoted as $\tau_i^{D}$ where $i \in \mathcal{I}_D$ (and $\mathcal{I}_D$ is the set of indices corresponding to data-dependent segments), we apply pose transformation to adapt to changes in object pose in the new scene.This transformation preserves the relative spatial relationship between the target pose and the object pose, ensuring consistent task execution across different scenes.We will discuss this in detail in \cref{second}.
\\\\\textbf{Replanning with VLM-Guided Constraints.}
For replanning segments $\tau_i^{R}$ ($i \in \mathcal{I}_R$), our framework generates executable trajectories through VLM-constrained optimization. A key innovation of this approach is the translation of VLMs' semantic understanding into kinematic constraints, achieved through a two-phase process. First, in \textbf{semantic keypoint extraction}, we extract task-critical points $\mathcal{K} = \{k_j\}_{j=1}^J$ from the environment using clip based feature alignment. For instance, in the Threading task, keypoints could include the needle tip and the tripod hole. Each keypoint $k_j \in \mathbb{R}^3$ pinpoints a semantically relevant location, such as object contact surfaces or edges, establishing spatial anchors for constraint formulation. Second, in \textbf{spatiotemporal constraint generation}, Gemini analyzes both the video dynamics and the RGB image of the environment, now annotated with the keypoints. For the Threading task, Gemini might generate a constraint such as: ``maintain the needle horizontally oriented while moving it towards the tripod hole". These generated constraints, reflecting the task's semantic requirements, guide the path planner in producing trajectories $\tau = \{\mathbf{T}_k\}$ by solving a constrained optimization problem.
The path planner then generates $\tau = \{\mathbf{T}_k\}$ by solving the following constrained optimization problem:

\begin{equation}
\label{eq:constrained_optimization}
\begin{aligned}
\min_{\{\mathbf{c}_i\}} \quad & \lambda_p J_p + \lambda_c J_c + \lambda_l J_l + \lambda_{ik} J_{ik} \\
\text{s.t.} \quad & \forall T_k \in \mathcal{T}_R: T_k \in \mathcal{C}_{VLM} \\
\end{aligned}
\end{equation}
where we define each cost term as:
\begin{itemize}
    \item $J_p = \sum_{k\in\mathcal{T}_R} \|\mathbf{T}_t\mathcal{K} - \mathcal{C}_{VLM}\|_{\Sigma}$ is the VLM semantic constraint cost.
    \item $J_c$ is the collision cost penalizing obstacle penetration using SDF distances.
    \item $J_l$ is the smoothness cost combining positional $\ell_2$-norm and rotational geodesic distance.
    \item $J_{ik}$ is the inverse kinematics (IK) loss, encouraging kinematic feasibility.
\end{itemize}

The path planning in HybridGen is guided by two key types of constraints: VLM-derived constraints and kinematic feasibility constraints. The VLM-derived constraints, informed by the VLM's temporal analysis of expert demonstrations, ensure the semantic validity of the generated trajectories, during replanning segments $\mathcal{T}_R$. Simultaneously, classical planning methods enforce kinematic feasibility, ensuring the robot's physical limitations are respected. This dual approach overcomes the ``semantic-mechanical disconnect" often encountered when relying solely on VLMs for robot control. To further enhance kinematic feasibility, especially in the generated trajectories, an inverse kinematics loss ($J_{ik}$) is incorporated into the optimization objective. For clarity and consistency, all constraint terms are represented using the notation $J$. The detailed formulation of the inverse kinematics loss and further elaborations will be provided in subsequent sections.
\subsection{Second augmentation}
\label{second}
In the second augmentation phase, we perform large-scale expansion on the dataset $D_{src}^{'}$ generated in the first stage. To balance effectiveness and efficiency, this stage optimizes each pose for scene adaptation without considering trajectory categorization. Our experimental tasks involve interactions between a grasped object $A$ and a target object $B$. Previous methods constrained subtask selection by maintaining consistency with the source dataset's grasping poses for object $A$, as they only considered end-effector poses relative to $B$. We address this limitation by centering our formulation around the grasped object $A$, explicitly modeling its relative pose with respect to $B$. This enables free subtask selection across all stages while fully utilizing the potential of the first-stage augmented dataset. 

To implement this, we propose the \textit{Nearest Grasp Object Relative to Target Object} selection strategy. Given a current grasp object pose $T^{G}_W \in SE(3)$ and target object pose $T^{O}_W \in SE(3)$, we compute the relative grasp pose as $T^{G}_O = (T^{O}_W)^{-1} T^{G}_W$. For each source demonstration with grasp pose $T^{G'}_W$ and target object pose $T^{O'}_W$, we calculate the source relative grasp pose $T^{G'}_{O'} = (T^{O'}_W)^{-1} T^{G'}_W$. 

The strategy selects top-$k$ candidates with minimal distances through:

\begin{equation}
    \mathcal{K} = \text{argtopk}_i \left[ d(T^{G}_O, T^{G'_i}_{O'_i}) \right]
\end{equation}

To adapt the source data to the new scene, we propose a key assumption that the relative pose between the grasped object and the target object in the current scene remains the same as in the source data:
\begin{equation}
\left(T_{W}^{O}\right)^{-1} T_{W}^{G} = \left(T_{W}^{O'}\right)^{-1} T_{W}^{G'}
\label{keyassumption}
\end{equation}
According to the Eq. \ref{keyassumption}, we calculate the pose of the end-effector in the new scene in the following way:
\begin{equation}
    T_W^E = \underbrace{T_W^O (T_{W'}^{O'})^{-1} T_{W'}^{G'}}_{T_W^G} T_G^E
\end{equation}
where $T^{E}_W$ represents the transformed end-effector pose and $T^{E'}_{W'}$ denotes the original end-effector pose from the source demonstration(derivation in Appendix \ref{Derivation}). This formulation preserves the relative motion between grasped and target objects while enabling flexible grasp pose selection across different demonstrations.This object-centric approach decouples subtask selection from the absolute grasping poses of the source demonstrations. Consequently, HybridGen achieves ``free subtask selection" across all stages of the task. This means that the system can now select and recombine subtasks more liberally, fully leveraging the diversity introduced in the first augmentation stage and generating a much richer and more varied dataset.

\subsection{Constraint Extract}
\label{constraint}
\textbf{Keypoints Extraction.}
\begin{figure}[htbp]  
    \centering 
  
    \begin{subfigure}[b]{0.225\textwidth} 
      \centering 
      \includegraphics[width=\textwidth]{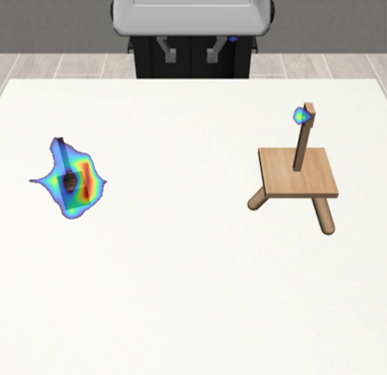} 
      \caption{Heatmap} 
      \label{fig:sub1} 
    \end{subfigure}
    \hfill 
    \begin{subfigure}[b]{0.225\textwidth} 
      \centering
      \includegraphics[width=\textwidth]{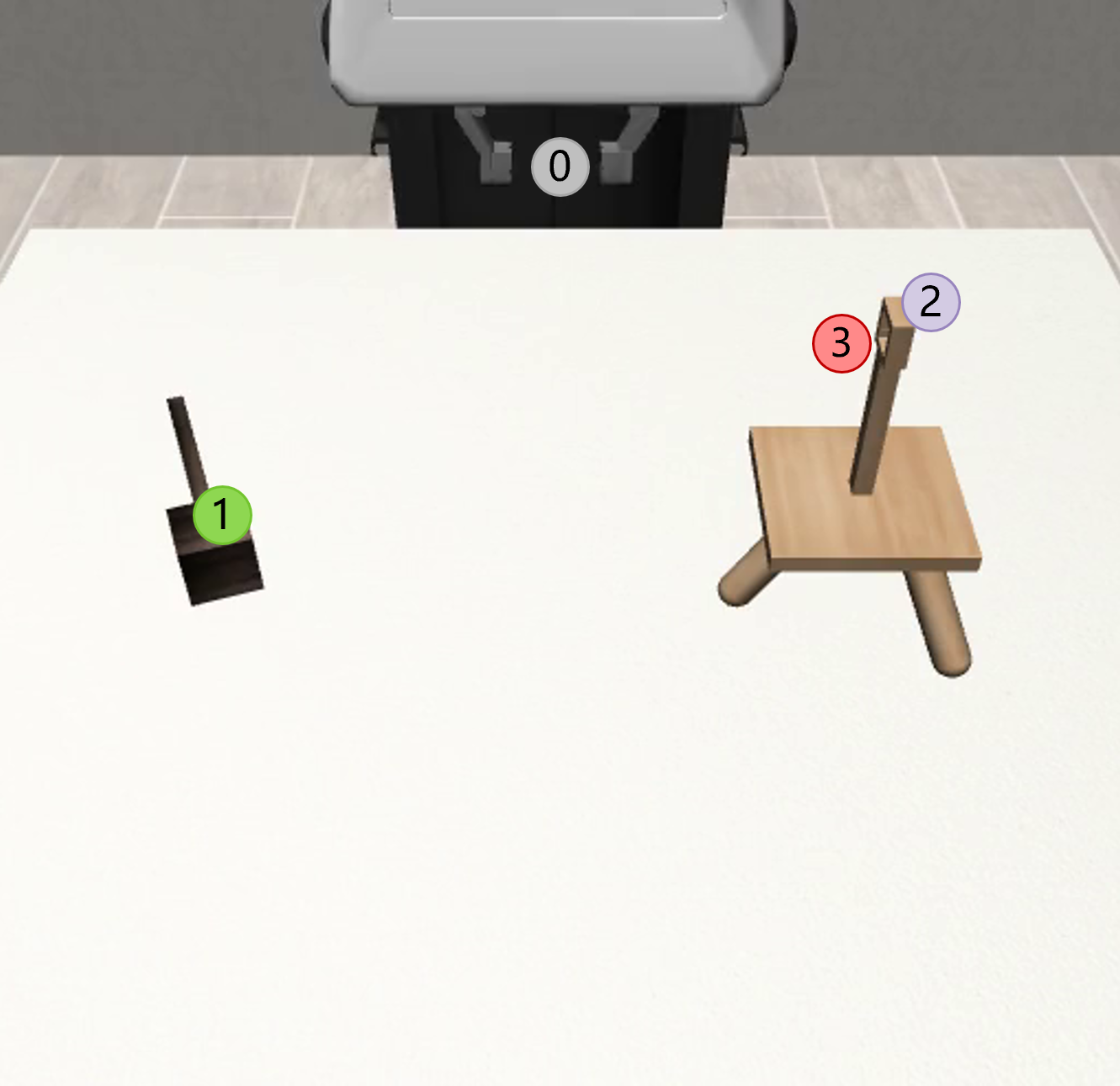} 
      \caption{Keypoints}
      \label{fig:sub2}
    \end{subfigure}
  
    \caption{Keypoint Extraction Process. (a) Heatmap generated by CLIP highlighting regions semantically relevant to the task description. Brighter areas indicate higher relevance. (b) Keypoints extracted from the heatmap, representing task-relevant spatial locations. The keypoint 0 is always the end-effector.} 
    \label{fig:main} 
  \end{figure}
Some existing works enable VLMs to guide robotic task execution in a fine-grained manner by representing manipulation tasks as a series of relational keypoint constraints \cite{huangReKepSpatioTemporalReasoning,zhouCodeasMonitorConstraintawareVisual2024}. However, these approaches extract keypoints solely based on images while neglecting task requirements, potentially introducing task-irrelevant keypoints. We propose a CLIP (Contrastive Language-Image Pre-training)-based constraint point extraction method that aims to extract semantically meaningful, task-relevant keypoints from RGB images using scene-specific task descriptions as textual prompts, which subsequently serve as constraints for path planning.  

Given an input RGB image \( I \in \mathbb{R}^{H \times W \times 3} \) and a scene task description text prompt \( T \), the text prompt \( T \) representing the scene task description is first encoded into a text feature vector \( F_{text} = \text{CLIP}_{\text{text}}(T) \) via CLIP's text encoder. Simultaneously, the input image \( I \) is processed through CLIP's visual encoder, \( \text{CLIP}_{\text{image}}(\cdot) \). Crucially, unlike methods that only extract a single global image embedding, we leverage CLIP to obtain dense, patch-level features. The image \( I \) is preprocessed and effectively divided into a grid of \( N \) patches, \( \{P_i\}_{i=1}^N \), where each patch \( P_i \) corresponds to a region of the input image. The visual encoder then produces a feature vector \( F_i \in \mathbb{R}^{D} \) for each patch, resulting in a set of patch embeddings \( \{F_i\}_{i=1}^N \). The dimensionality \( D \) of the patch embeddings depends on the specific CLIP model architecture(in our case, we use ViT \cite{dosovitskiy2020image}).
These patch embeddings, which represent localized regions of the image, are then used to compute a task-relevant feature response map. Specifically, the patch embeddings are first reshaped into a 2D feature map \( F_{\text{map}} \in \mathbb{R}^{H' \times W' \times D} \), where \( H' \) and \( W' \) represent the spatial dimensions of the patch grid. This reshaping reflects the spatial arrangement of the patches within the original image.

After obtaining the feature response map \( S \), we follow the methodology in Rekep \cite{huangReKepSpatioTemporalReasoning} to perform K-means clustering on high-response regions, identifying representative areas with elevated response values. Cluster centers are treated as candidate constraint points in the feature space. The 3D spatial points corresponding to the highest feature response values nearest to cluster centers are selected as candidate constraint points. Finally, we filter out candidates outside the workspace and apply post-processing steps including Mean Shift clustering-based merging to derive the final set of constraint points, which are annotated on the scene's RGB image and mapped to 3D coordinates.  
\\\\\textbf{Constraint Proposal.} 
We employ Gemini \cite{geminiteamGeminiFamilyHighly2024} to generate keypoint-described constraint conditions \( c_{path} \) for each sub-stage derived from task decomposition. Unlike prior approaches that primarily relied on textual task descriptions and static RGB images for constraint generation, our method incorporates a richer multimodal input to Gemini. This input comprises: (1) the textual task description, providing semantic context; (2) RGB images of the environment, annotated with extracted keypoints representing task-relevant spatial locations; and crucially, (3) video recordings of the source demonstration replays (see Appendix \ref{query} for the prompt). This inclusion of video dynamics as input enables Gemini to capture temporal motion patterns inherent in successful demonstrations, leading to the generation of more accurate and contextually relevant constraints. For each subtask, we pre-annotate temporal segments within the demonstration video. This temporal segmentation allows Gemini to precisely generate constraints tailored to each individual subtask rather than for broader, less specific phases. Notably, the proposed constraints are designed to span the entirety of each subtask execution, ensuring consistency and coherence throughout the subtask's execution process, rather than being limited to only the explicitly plannable segments.

%% file: sec/4_experiments.tex
\section{Experiments}
\begin{figure*}[ht] 
    \centering
    \begin{subfigure}[b]{0.13\textwidth}
      \includegraphics[width=\textwidth]{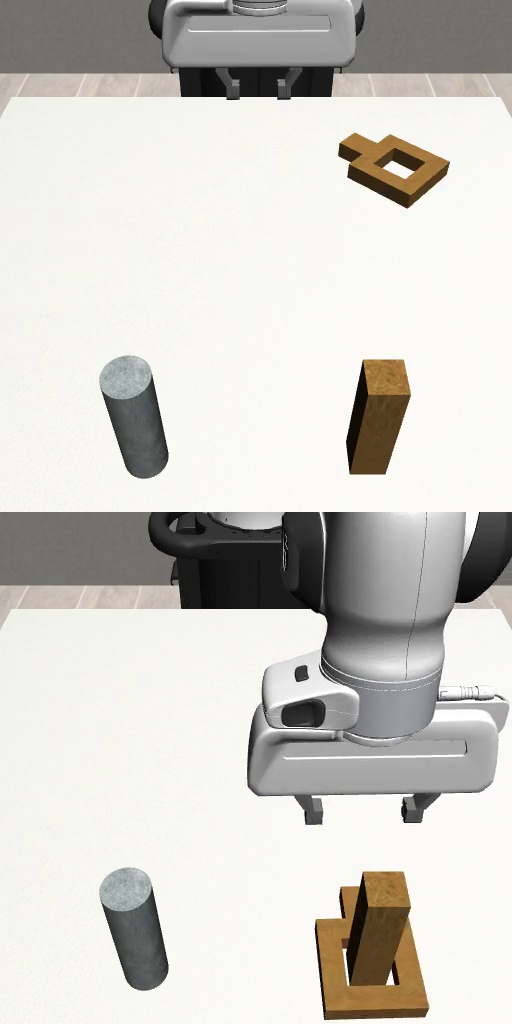} 
      \caption{Square}
      \label{tasks:a}
    \end{subfigure}
    \hfill 
    \begin{subfigure}[b]{0.13\textwidth}
      \includegraphics[width=\textwidth]{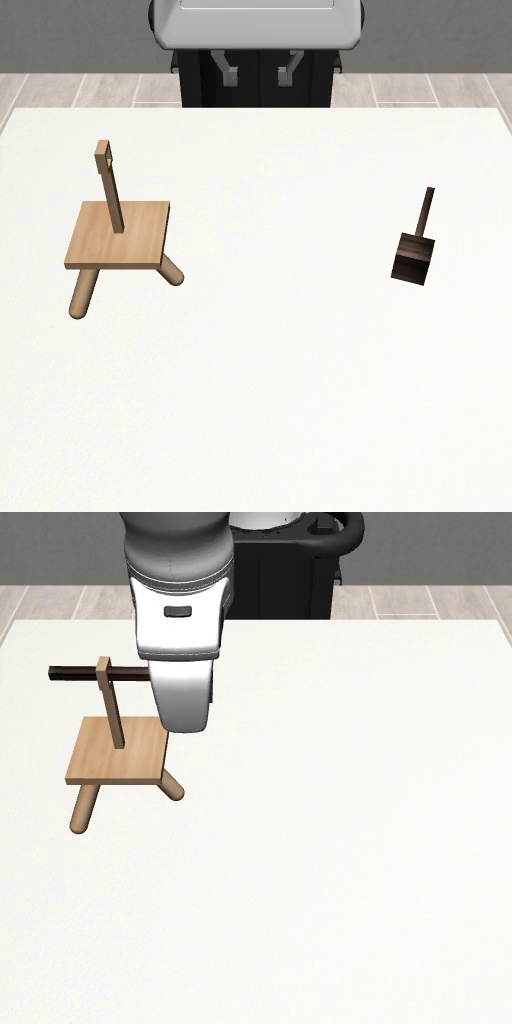}
      \caption{Threading}
      \label{tasks:b}
    \end{subfigure}
    \hfill
    \begin{subfigure}[b]{0.13\textwidth}
      \includegraphics[width=\textwidth]{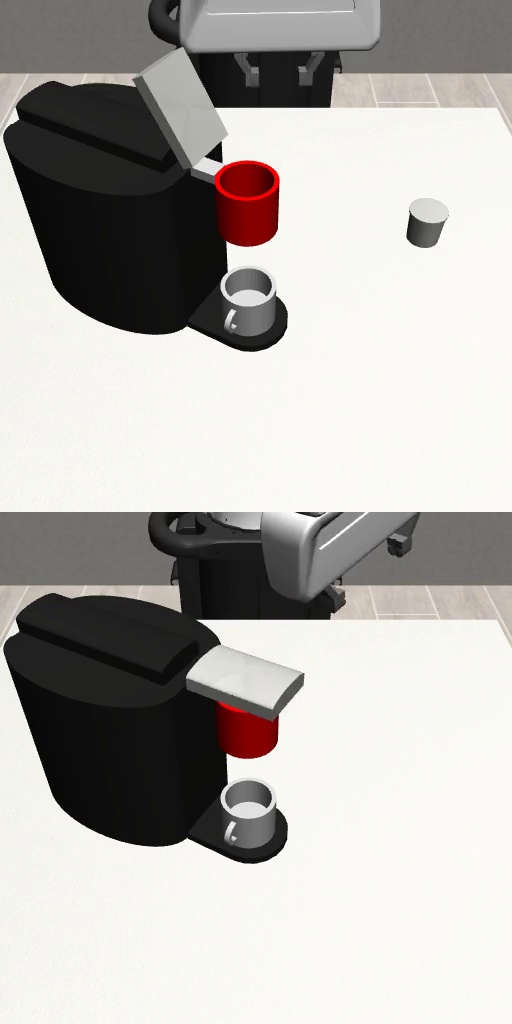}
      \caption{Coffee}
      \label{fig:sub3}
    \end{subfigure}
        \hfill
    \begin{subfigure}[b]{0.13\textwidth}
      \includegraphics[width=\textwidth]{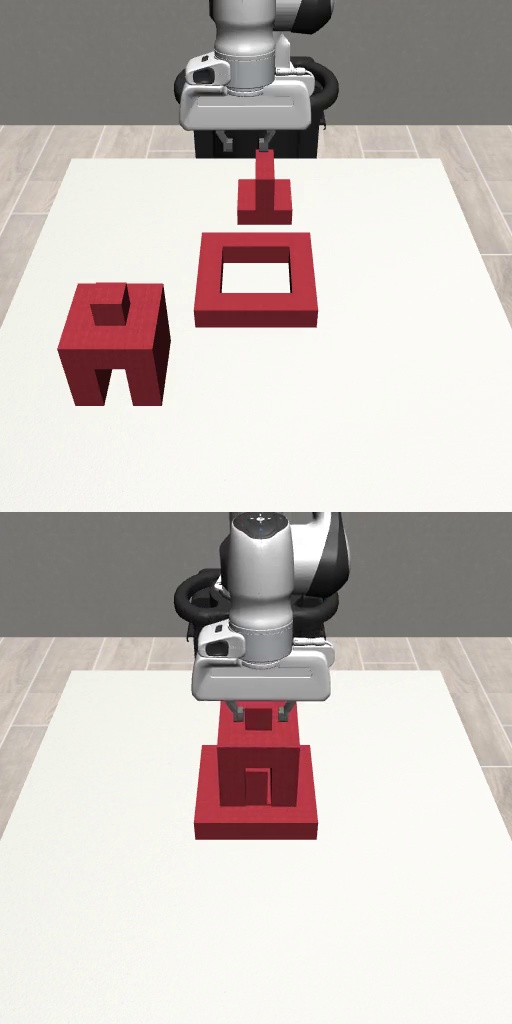}
      \caption{Assembly}
      \label{fig:sub4}
    \end{subfigure}
        \hfill
    \begin{subfigure}[b]{0.13\textwidth}
      \includegraphics[width=\textwidth]{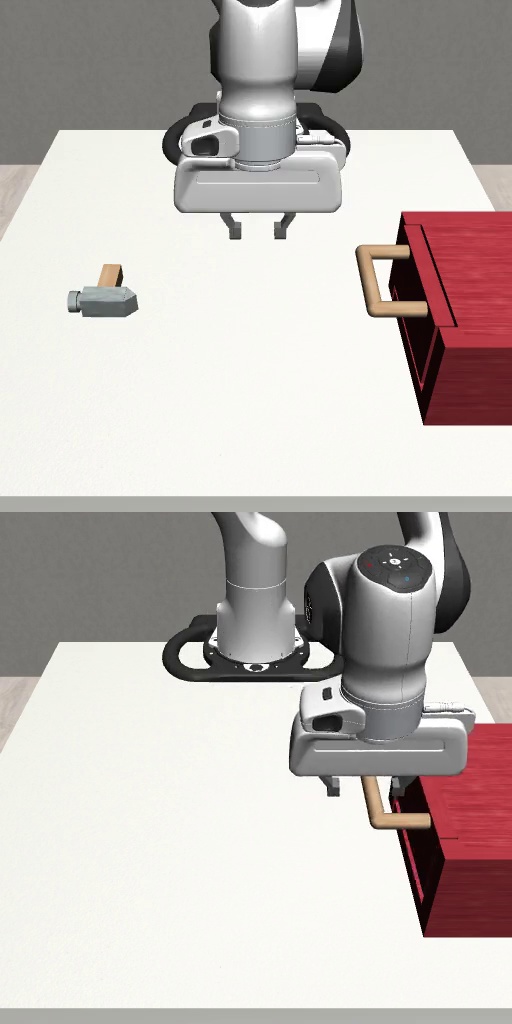}
      \caption{Hammer}
      \label{fig:sub5}
    \end{subfigure}
        \hfill
    \begin{subfigure}[b]{0.13\textwidth}
      \includegraphics[width=\textwidth]{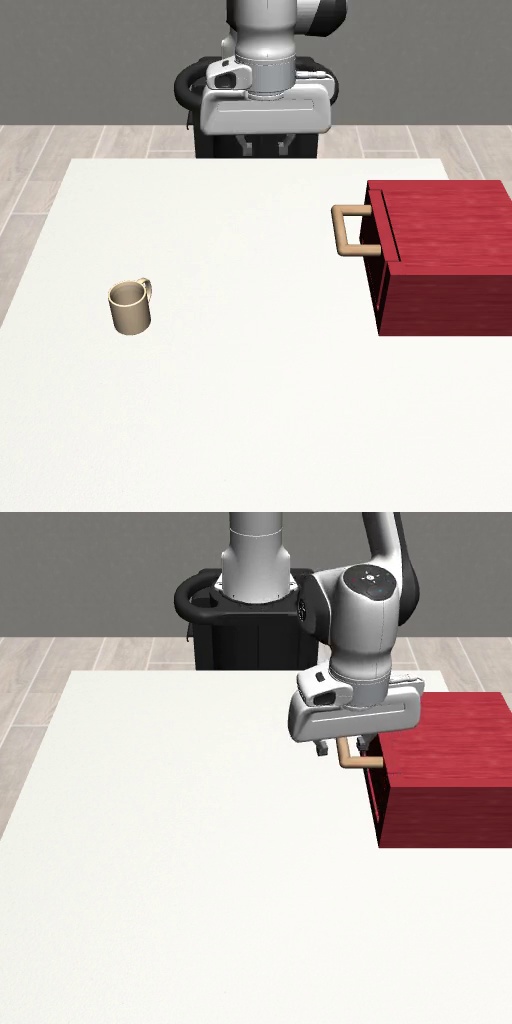}
      \caption{Mug}
      \label{fig:sub6}
    \end{subfigure}
        \hfill
    \begin{subfigure}[b]{0.13\textwidth}
      \includegraphics[width=\textwidth]{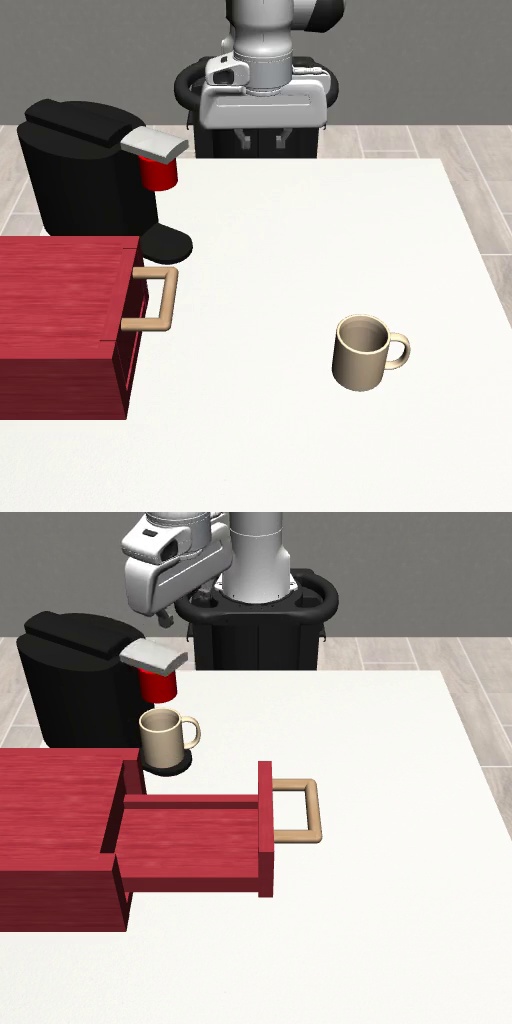}
      \caption{Coffee Prep}
      \label{fig:sub7}
    \end{subfigure}
  
    \caption{\textbf{Tasks}. We evaluate the HybridGen framework on seven manipulation tasks and their variants. Each task involves interaction between a grasped object and a target object. Tasks (a-c) require fine-grained manipulation and precise control. Tasks (d-g) are multi-stage tasks that test the agent's ability to plan over long horizons.}
    \label{fig:tasks}
  \end{figure*}
We conduct comprehensive experiments to validate three key claims: (1) Hybridgen demonstrates strong performance across diverse manipulation tasks, (2) The generated data exhibits broad utility for training various imitation learning algorithms, and (3) Our proposed VLM-based two-stage generation strategy and subtask selection mechanism significantly contribute to system effectiveness.
\subsection{Experimental Setup}
\textbf{Task Configuration.}
Tasks are evaluated across multiple difficulty variants. We test HybridGen on the seven tasks (18 total variants) shown in Fig. \ref{fig:main}. These can be broadly categorized into: Fine-grained Manipulation and Precise Control, encompassing Square, Threading, and Coffee. These tasks require the agent to execute precise movements and delicate interactions between the grasped object and the target. The second category is Long-Horizon Task, which includes Piece Assembly, Hammer Cleanup, Mug Cleanup, and Coffee Prep. These tasks involve multiple stages and test the agent's ability to plan and execute actions over extended periods to achieve the final goal. Each task configuration includes $D_0$ with fixed object placements upon environment reset, $D_1$ with broader object distributions than $D_0$, and where applicable, $D_2$ featuring even greater diversity following Mmicgen protocols(full details in Appendix \ref{taskconfig}). This setup allows us to evaluate the robustness of the learned policies to variations in initial conditions.
\\\\\textbf{Data Generation.}
We begin with 10 human demonstrations from the Robomimic Square PH dataset [7] for each task. Our two-stage augmentation first expands this to 50 demonstrations via VLM-guided interactions, and then to 1,000 demonstrations. Each generated demonstrations is executed in the simulation environment; only successful demonstrations are kept. During subtask selection (Sec. \ref{second}), we choose from the top-k candidates, with k = 3 in our experiments.
\\\\\textbf{Path Planning.}
Path planning stage employs a constrained optimization approach, with the specific objective function as shown in Eq. \ref{eq:constrained_optimization}, encompassing the following cost terms: VLM semantic constraint cost ($J_p$), collision cost ($J_c$), smoothness cost ($J_i$), and inverse kinematics loss ($J_{iK}$). To effectively balance the contributions of each cost term, we set the following weighting parameters: VLM semantic constraint cost weight $\lambda_p = 100$, collision cost weight $\lambda_c = 1$, smoothness cost weight $\lambda_i = 0.1$, and inverse kinematics loss weight $\lambda_{ik} = 20$. For the implementation of inverse kinematics constraints, we integrated IKFast \cite{diankov2010automated}, an efficient analytical inverse kinematics solver, to ensure that the generated demonstrations are kinematically feasible.
\\\\\textbf{Imitation Learning.}
We trained BC-RNN implementation from the Robomimic framework on the data generated by HybridGen and compared it with Mimicgen,maintaining the training parameters consistent with those used in MimicGen for a fair comparison. We additionally selected BC-Transformer and Diffusion Policy algorithms to test the algorithmic robustness of HybridGen, using Robomimic's default parameters for all parameters (full details in Appendix \ref{training}).

\subsection{Baseline Comparisons}

\begin{table}
  \centering
  \begin{tabular}{lcc}
  \toprule
  \bf{Task Variant} & \bf{Mimicgen} & \bf{HybridGen} \\
  \midrule
  Square $D_0$ & $90.7\pm1.9$ & $\mathbf{100.0\pm0.0}$ \\
  Square $D_1$ & $73.3\pm3.4$  & $\mathbf{74.7\pm3.4}$ \\
  Square $D_2$ & $49.3\pm2.5$ & $\mathbf{57.3\pm2.5}$ \\
  \midrule
  Threading $D_0$ & $98.0\pm1.6$  & $\mathbf{100.0\pm0.0}$ \\
  Threading $D_1$ & $60.7\pm2.5$  & $\mathbf{63.3\pm0.9}$ \\
  Threading $D_2$ & $38.0\pm3.3$  & $\mathbf{46.0\pm1.6}$ \\
  \midrule
  Coffee $D_0$ &$ \mathbf{100\pm0.0}$  & $\mathbf{100.0\pm0.0}$ \\
  Coffee $D_1$ & $90.7\pm2.5$  & $\mathbf{94.7\pm3.8}$ \\
  Coffee $D_2$ & $\mathbf{77.3\pm0.9}$  & $74.7\pm3.4$ \\
  \midrule
  Piece Assembly $D_0$ & $82.0\pm1.6$  & $\mathbf{85.3\pm0.9}$ \\
  Piece Assembly $D_1$ & $\mathbf{62.7\pm2.5}$  & $62.0\pm4.3$ \\
  Piece Assembly $D_2$ & $13.3\pm3.8$  & $\mathbf{16.7\pm2.5}$ \\
  \midrule
  Hammer Cleanup $D_0$ & $\mathbf{100.0\pm0.0}$  & $\mathbf{100.0\pm0.0}$ \\
  Hammer Cleanup $D_1$ & $62.7\pm4.7$  & $\mathbf{80.7\pm0.9}$ \\
  \midrule
  Mug Cleanup $D_0$ & $80.0\pm4.9$  & $\mathbf{88.7\pm2.5}$\\
  Mug Cleanup $D_1$ & $64.0\pm3.3$  & $\mathbf{86.0\pm2.8}$ \\
  \midrule
  Coffee Prep $D_0$ & $\mathbf{97.3\pm0.9}$  & $84.7\pm2.5$ \\
  Coffee Prep $D_1$ & $42.0\pm0.0$  & $\mathbf{56.7\pm0.9}$ \\
  \midrule
  \bf{Average} & 71.2 & \bf{76.2} \\
  \bottomrule
  \end{tabular}
  \caption{\textbf{Agent Performance on HybridGen Datasets}. Success rates (3 seeds) and standard deviations for agents trained using data generated by HybridGen and Mimicgen across seven different manipulation tasks and their difficulty variants ($D_0$, $D_1$, $D_2$). HybridGen demonstrates consistently superior performance compared to Mimicgen across a wide range of tasks and difficulty levels, highlighting the effectiveness of our approach in generating high-quality and diverse training data for imitation learning.}
  \label{baseline}
\end{table}

To validate HybridGen's performance advantage, we conducted comparisons against the Mimicgen baseline. \ref{baseline} meticulously presents the performance of HybridGen and Mimicgen across seven diverse manipulation tasks and their difficulty variants. These tasks span from fine-grained manipulations (Square, Threading, Coffee) to multi-stage tasks (Piece Assembly, Hammer Cleanup, Mug Cleanup, Coffee Prep), with difficulty levels increasing from $D_0$ (fixed object placements) to $D_2$ (higher object distribution diversity).

As clearly shown in Table \ref{baseline}, HybridGen consistently outperforms Mimicgen across most tasks with a 5\% average improvement. Notably, HybridGen frequently achieves near-perfect success rates (100\%) in lower difficulty $D_0$ tasks. Particularly in the most challenging variant of each task, HybridGen demonstrates exceptional performance, achieving an average success rate of 59.7\%, significantly surpassing Mimicgen's 49.5\%.

These results strongly demonstrate HybridGen's superior performance compared to Mimicgen across a range of manipulation tasks, signifying its effectiveness as an advanced data generation framework for enhancing imitation learning algorithm performance and generalization.
\subsection{Algorithmic Robustness}
\begin{figure}
  \centering
   \includegraphics[width=1\linewidth]{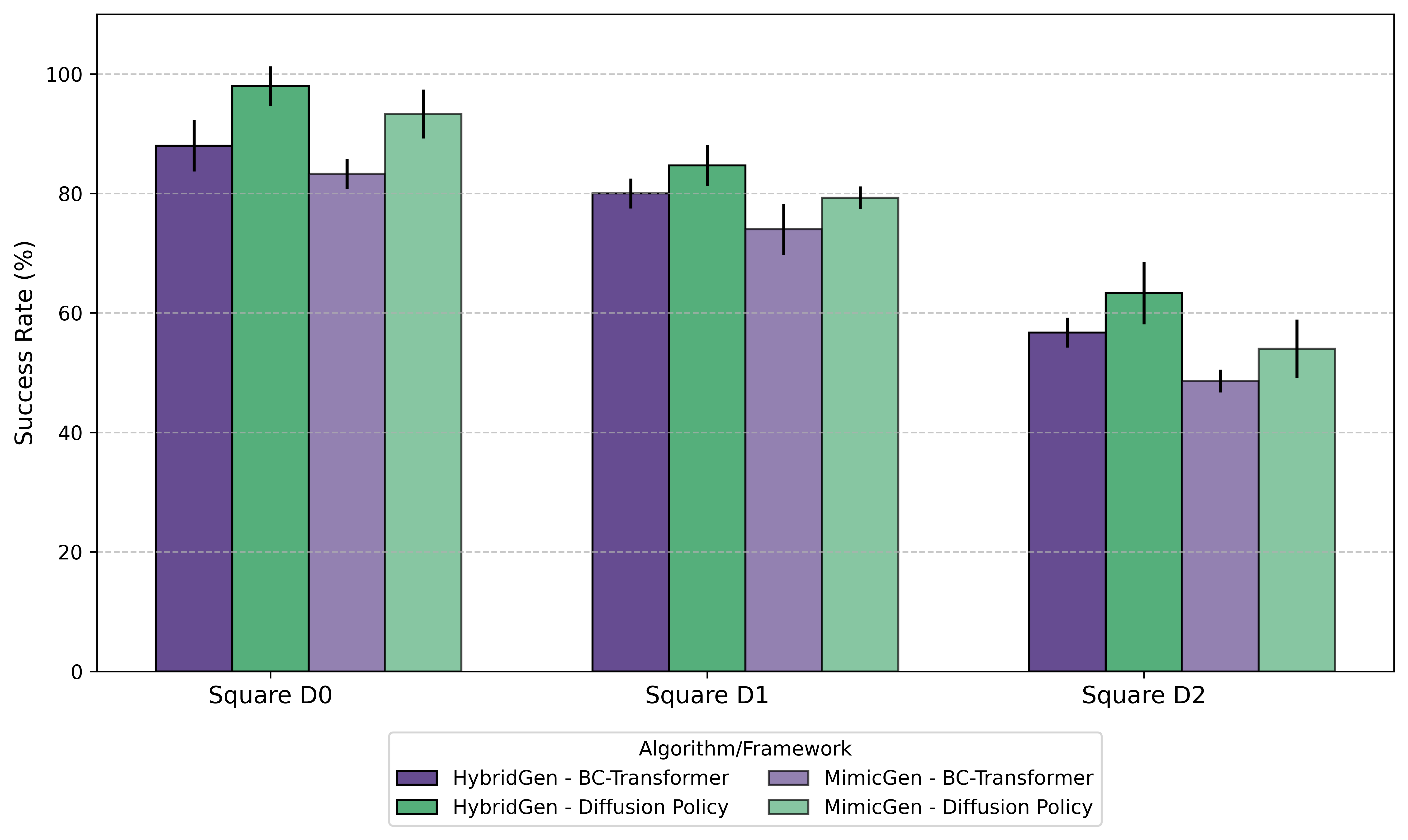}
   \caption{\textbf{Effectiveness across algorithms of HybridGen}. Success rates (3 seeds) comparison between HybridGen and MimicGen on Square task variants ($D_0$, $D_1$, $D_2$) using BC-Transformer and Diffusion Policy architectures. HybridGen demonstrates consistent performance advantages across fundamentally different policy learning paradigms, validating its ability to generate universally useful demonstration data for diverse imitation learning approaches.}
   \label{fig:compare}
\end{figure}

To validate the broad applicability of the data generated by HybridGen, we evaluated its effectiveness across different imitation learning algorithms. We selected two state-of-the-art approaches representing distinct policy learning paradigms: BC-Transformer and Diffusion Policy. BC-Transformer , a Transformer-based behavior cloning approach, leverages self-attention mechanisms to model long-horizon task dependencies. Its architecture is particularly well-suited for capturing sequential patterns present in demonstration data, making it highly effective for multi-stage tasks requiring temporal coherence. Diffusion Policy, on the other hand, employs a diffusion probabilistic model to iteratively refine action sequences through a denoising process. This method inherently possesses robustness to stochasticity and environmental dynamics, which is highly beneficial for tasks with inherent uncertainty. To verify the versatility of HybridGen data across different algorithms, and thus demonstrate its algorithmic robustness, we subsequently trained agents using both BC-Transformer and Diffusion Policy algorithms, each trained separately on data generated by HybridGen and MimicGen, and then compared the performance of the trained agents on various variants of the Square task ($D_0$, $D_1$, and $D_2$).
Fig. \ref{fig:compare}compellingly showcases the effectiveness across algorithms of HybridGen data for imitation learning. Agents trained with HybridGen-generated data consistently achieve significantly higher success rates compared to those trained with MimicGen data, a trend observed across varying difficulty levels and when employing both BC-Transformer and Diffusion Policy architectures. HybridGen data demonstrably facilitates enhanced learning and improved generalization, particularly in complex scenarios. Crucially, HybridGen consistently surpasses MimicGen's performance across all difficulty levels and with both learning algorithms. This unequivocally underscores the effectiveness of our data generation framework in enhancing imitation learning performance and generalization, especially under challenging conditions,confirming its broad applicability across diverse imitation learning algorithms.

\begin{table}
  \centering
  \begin{tabular}{cccccc}
    \toprule
    \multirow{2}{*}{\#} & \multirow{2}{*}{VLM} & \multirow{2}{*}{GRT} & \multicolumn{3}{c}{Tasks} \\
  
     &  &  & Square & Threading & \\
    \midrule
    (a) & \textcolor{blue}{\checkmark} & \textcolor{blue}{\checkmark} & \bf{58.7} & \bf{46.7} \\
    (b) & \textcolor{red}{\ding{55}} & \textcolor{blue}{\checkmark} & 51.3 & 44.7 \\
    (c) & \textcolor{blue}{\checkmark} & \textcolor{red}{\ding{55}} & 54.7 &  42.0\\
    (d) & \textcolor{red}{\ding{55}} & \textcolor{red}{\ding{55}} & 49.3 & 38.0 \\
    \bottomrule
  \end{tabular}
  \caption{\textbf{Ablation Study}. Success rates (3 seeds) for different ablation conditions of HybridGen on the Square $D_2$ and Threading $D_2$ tasks. The columns represent the presence or absence of VLM planning (``VLM") and \textit{Nearest Grasp Object Relative to Target Object subtask selection strategy}(``GRT"). Condition (a) represents the full HybridGen system. Conditions (b), (c), and (d) systematically disable VLM and/or GRT to evaluate their individual contributions. The results demonstrate that both VLM planning and GRT subtask selection are crucial for achieving optimal performance.}
  \label{ablation}
  \end{table}

\subsection{Ablation Studies}
To validate the effectiveness of our framework's core components, we conducted a series of ablation studies on the Square $D_2$ and Threading $D_2$, focusing on the VLM planning strategy and the subtask selection strategy (GRT). As shown in Table \ref{ablation}, we systematically disabled these components to assess their individual contributions. Condition (a) represents the full HybridGen system with both VLM planning and GRT-based subtask selection enabled. Condition (b) disables the VLM, relying solely on linear interpolation for trajectory generation while retaining the GRT selection strategy. Condition (c) utilizes the VLM for planning but replaces GRT with a random subtask selection mechanism. Finally, condition (d) disables both the VLM and GRT, representing a baseline approach.

The results clearly demonstrate the importance of both components. Comparing (a) and (b), we observe a performance drop on both the Square (58.7\% to 51.3\%) and Threading (46.7\% to 44.7\%) tasks when the VLM is removed. This highlights the VLM's crucial role in generating high-quality, task-relevant trajectories that go beyond simple interpolation. Similarly, comparing (a) and (c) reveals the benefit of the GRT-based subtask selection. While using the VLM alone (c) improves upon the baseline (d), the full system (a) achieves superior performance. This indicates that GRT effectively identifies and prioritizes the most beneficial subtasks for policy learning, leading to improved overall task success. Finally, the baseline condition (d), lacking both VLM planning and informed subtask selection, exhibits the lowest performance, reinforcing the synergistic contribution of both components to HybridGen's effectiveness.

%% file: sec/5_conclusion.tex
\section{Conclusion}
We introduced HybridGen, an effective and practical system to generating scalable imitation learning data. By integrating VLMs with a novel two-stage hybrid planning framework, HybridGen significantly enhances data diversity and quality from limited demonstrations. Evaluations across a range of robotic tasks demonstrate substantial performance gains and improved generalization compared to existing methods. This work underscores the potential of VLM-guided data augmentation to advance the field of imitation learning and enable more robust robotic manipulation capabilities.

%% file: sec/X_suppl.tex
\clearpage
\appendix
\setcounter{page}{1}
\maketitlesupplementary
\renewcommand{\thefigure}{\Alph{section}.\arabic{figure}}
\setcounter{figure}{0} 
\setcounter{page}{1}

\section{Policy Training Details}
\label{training}
We elaborate on the specifics of the policy training procedure, which was conducted via imitation learning.  Several design choices are consistent with the robomimic work.

\subsection*{Observation Spaces}

Following the methodology in robomimic\cite{mandlekar2022matters}, we train our policies using two distinct observation spaces: ``low-dim and ``image. Both observation types incorporate end effector poses and gripper finger positions.  Specifically, the ``image'' observation modality encompasses camera inputs from a front-view camera and a wrist-view camera.  All tasks utilize images with a resolution of 84x84, except for real-world tasks such as (Stack, Coffee), which employ a higher resolution of 120x160. For agents trained with ``image'' observations, we apply pixel shift randomization, and randomly shift image pixels by up to 10\% of each dimension every time an observation is presented to the agent.

\subsection*{Training Hyperparameters}
 
\subsubsection*{BC-RNN}

For BC-RNN, we utilized the Adam optimizer with an initial learning rate of $1e-4$. A multi-step learning rate decay schedule was employed, reducing the learning rate by a factor of 0.1 at specified epochs.  L2 regularization was not applied. The loss function consisted of an L2 loss term with a weight of 1.0. To model action distributions, we employed a Gaussian Mixture Model (GMM) with 5 modes, a minimum standard deviation of $1e-4$, and softplus activation for the standard deviation. Low-noise evaluation was enabled during inference.  The RNN architecture consisted of an LSTM with 2 layers and a hidden dimension of 1000.  The RNN horizon was set to 10, and the model operated in an open-loop fashion. We trained with a batch size of 16 for a total of 1000 epochs.

\subsubsection*{BC-Transformer}

For BC-Transformer, we utilized the AdamW optimizer with an initial learning rate of $1e-4$. A linear learning rate decay schedule was employed, reducing the learning rate by a factor of 0.1 after every 100 epochs.  We applied L2 regularization with a weight of 0.01. The loss function consisted of an L2 loss term with a weight of 1.0.  To model action distributions, we employed a Gaussian Mixture Model (GMM) with 5 modes, a minimum standard deviation of $1e-4$, and softplus activation for the standard deviation. Low-noise evaluation was enabled during inference. The transformer architecture consisted of 6 layers, an embedding dimension of 512, and 8 attention heads.  We trained with a batch size of 100 for a total of 1000 epochs.

\subsubsection*{Diffusion Policy}

For Diffusion Policy, we used an AdamW optimizer with a constant learning rate of $1e-4$. L2 regularization was not applied. The model horizon was configured with an observation horizon of 2, an action horizon of 8, and a prediction horizon of 8.  The UNet architecture for diffusion modeling included a diffusion step embedding dimension of 256, downsampling dimensions of [256, 512, 1024], a kernel size of 5, and 8 groups for group normalization. Exponential Moving Average (EMA) was enabled with a power of 0.75.  We used a DDPM setup with 100 training and 100 inference timesteps, a squared cosine beta schedule with capping (v2), clip sampling, and epsilon prediction type.  The batch size was set to 256, and training was conducted for 1000 epochs.

\subsection*{Policy Evaluation}

We assess policies on simulation tasks by performing 50 rollouts per agent checkpoint during training. We then report the maximum success rate achieved by each agent across 3 different random seeds.  For real-world tasks, due to the time-intensive nature of policy evaluation on physical robots, we evaluate only the last policy checkpoint generated during training, and assess its performance over 50 episodes.

\subsection*{Hardware}

Each data generation and training run was executed on a machine equipped with dual NVIDIA GeForce RTX 4090 GPUs, an Intel Core i9-14900K CPU, and 130GB of RAM. Each RTX 4090 GPU has 24GB of dedicated memory.  Disk storage was 6TB.

\section{Simulation Tasks}
\label{taskconfig}
\begin{figure*}[ht] 
    \centering
    \begin{subfigure}[b]{0.13\textwidth} 
      \includegraphics[width=\textwidth]{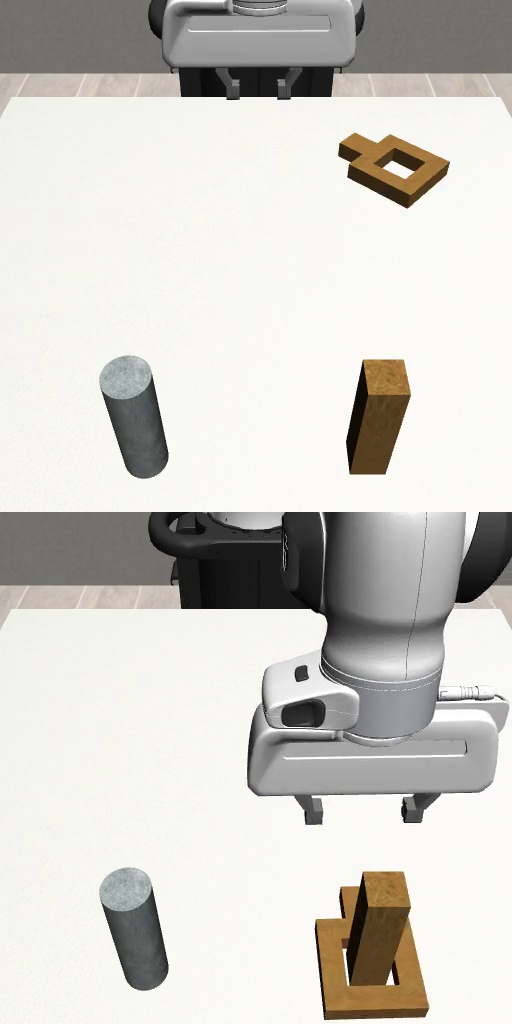}
      \caption{Square}
      \label{tasks:sa}
    \end{subfigure}
    \hfill 
    \begin{subfigure}[b]{0.13\textwidth}
      \includegraphics[width=\textwidth]{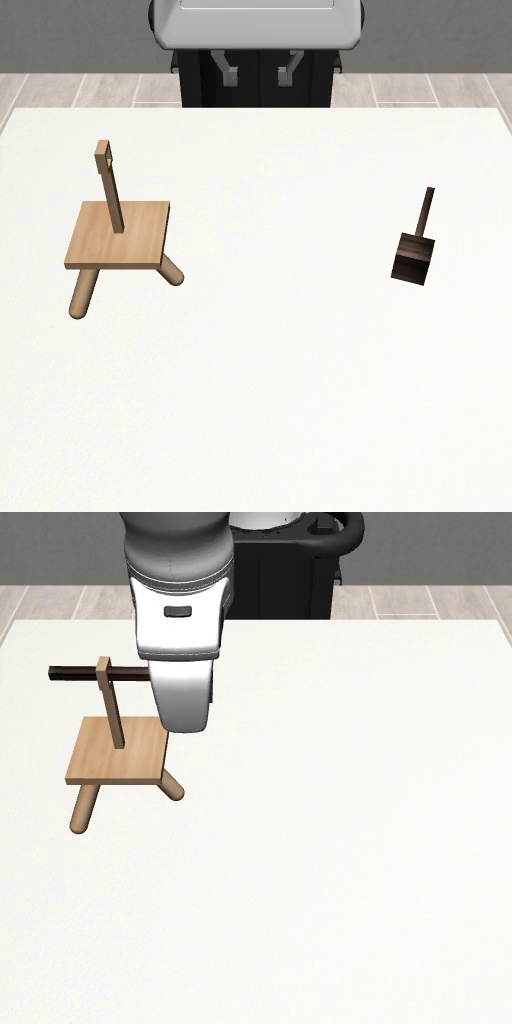}
      \caption{Threading}
      \label{tasks:sb}
    \end{subfigure}
    \hfill
    \begin{subfigure}[b]{0.13\textwidth}
      \includegraphics[width=\textwidth]{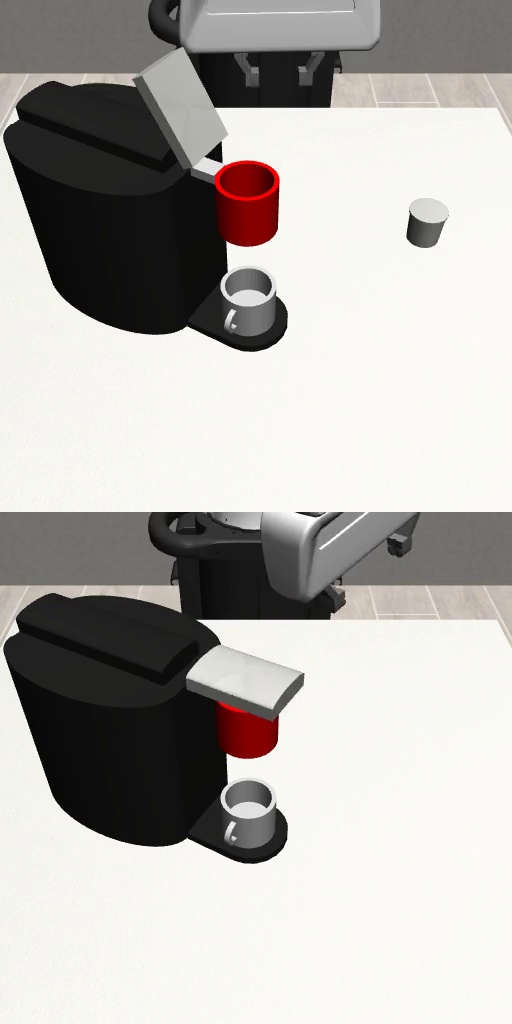}
      \caption{Coffee}
      \label{fig:subs3}
    \end{subfigure}
        \hfill
    \begin{subfigure}[b]{0.13\textwidth}
      \includegraphics[width=\textwidth]{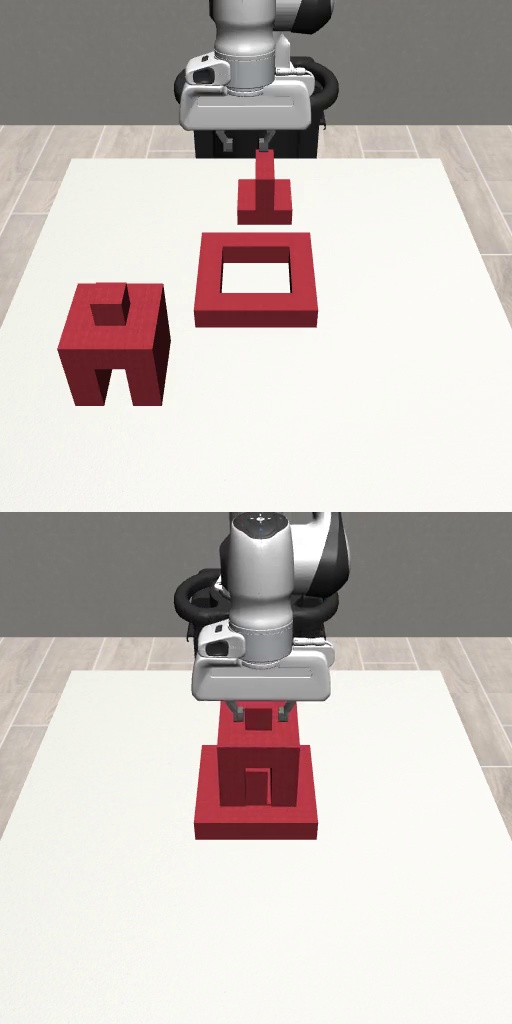}
      \caption{Assembly}
      \label{fig:subs4}
    \end{subfigure}
        \hfill
    \begin{subfigure}[b]{0.13\textwidth}
      \includegraphics[width=\textwidth]{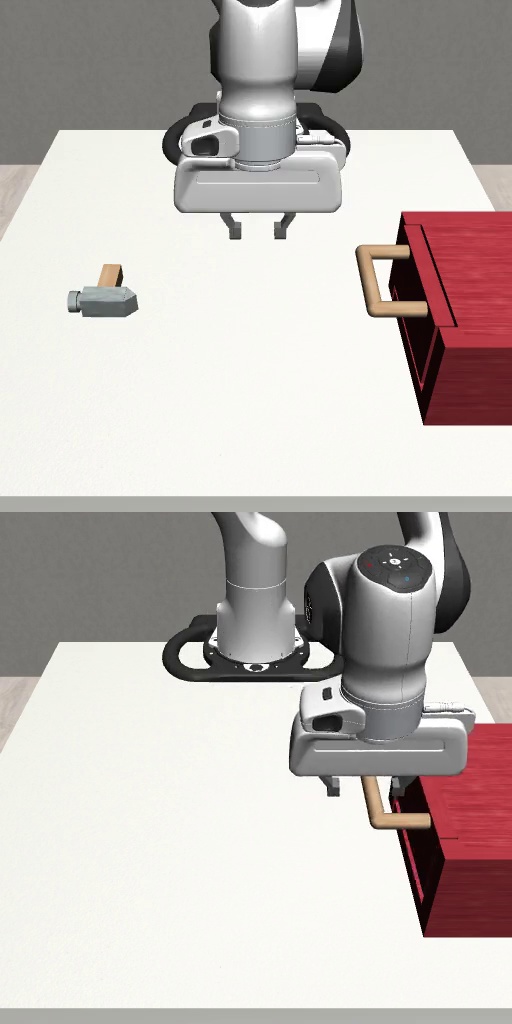}
      \caption{Hammer}
      \label{fig:subs5}
    \end{subfigure}
        \hfill
    \begin{subfigure}[b]{0.13\textwidth}
      \includegraphics[width=\textwidth]{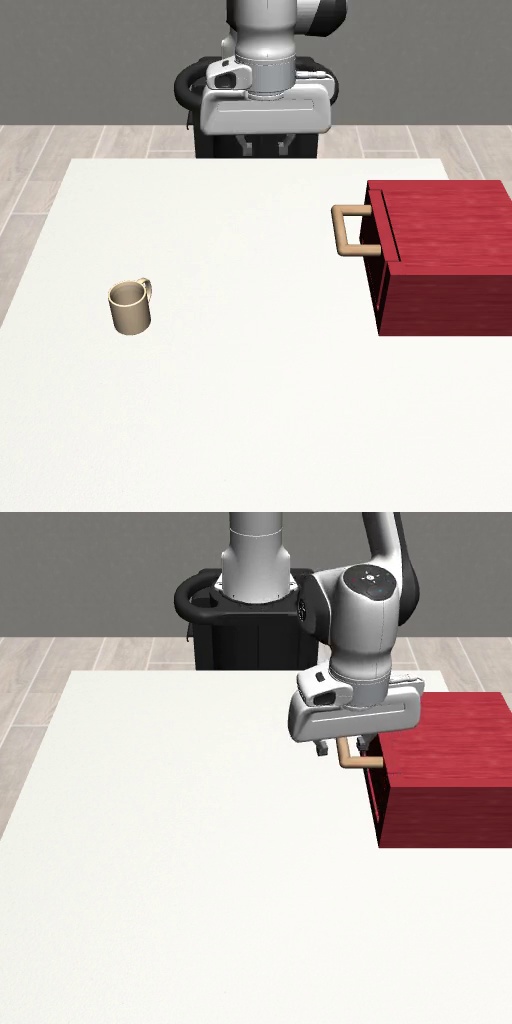}
      \caption{Mug}
      \label{fig:subs6}
    \end{subfigure}
        \hfill
    \begin{subfigure}[b]{0.13\textwidth}
      \includegraphics[width=\textwidth]{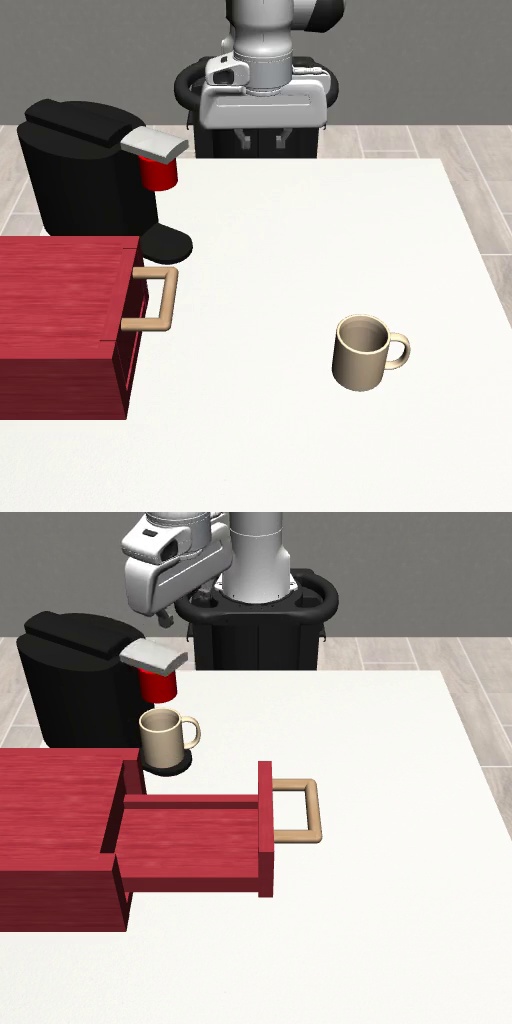}
      \caption{Coffee Prep}
      \label{fig:subs7}
    \end{subfigure}
  
    \caption{\textbf{Tasks}.Each task involves interaction between a grasped object and a target object. Tasks (a-c) require fine-grained manipulation and precise control. Tasks (d-g) are multi-stage tasks that test the agent's ability to plan over long horizons.}
    \label{sup:tasks}
  \end{figure*}

\textbf{Square Task}
\textbf{Goal:} Pick and place a square nut onto a peg.

\begin{itemize}
\item \textbf{D0 :} Fixed peg, nut initialized in a small region with random top-down rotation.
\item \textbf{D1 :} Peg and nut initialized in larger regions, fixed peg rotation, random nut top-down rotation.
\item \textbf{D2 :} Peg and nut initialized in larger regions, random peg and nut top-down rotation.
\end{itemize}

\textbf{Threading Task}
\textbf{Goal:} Pick a needle and thread it through a hole on a tripod.

\begin{itemize}
\item \textbf{D0 :} Fixed tripod, needle initialized in a modest region with limited top-down rotation variation.
\item \textbf{D1 :} Tripod and needle initialized in large regions (left and right of table), significant top-down rotation variation for both.
\item \textbf{D2 :} Similar to D1, but tripod and needle initialization positions reversed.
\end{itemize}

\textbf{Coffee Task}
\textbf{Goal:} Pick a coffee pod, insert it into a coffee machine, and close the machine hinge.

\begin{itemize}
\item \textbf{D0 :} Fixed machine, pod initialized in a small region.
\item \textbf{D1 :} Machine and pod initialized in larger regions (left and right of table), machine with top-down rotation variation.
\item \textbf{D2 :} Similar to D1, but machine and pod initialization positions reversed.
\end{itemize}

\textbf{Three Piece Assembly Task}
\textbf{Goal:} Assemble a three-piece structure by sequentially picking and inserting pieces.

\begin{itemize}
\item \textbf{D0 :} Fixed base, pieces initialized around base with fixed rotation.
\item \textbf{D1 :} All pieces initialized in a larger workspace region with fixed rotation.
\item \textbf{D2 :} All pieces initialized with rotation variation (base with smaller, pieces with larger variation).
\end{itemize}

\textbf{Nut Assembly Task}
\textbf{Goal:} Place both a square nut and a round nut onto two different pegs.

\begin{itemize}
\item \textbf{D0 :} Each nut initialized in a small box with random top-down rotation.
\item \textbf{D1 :} Nuts initialized in a large box with random top-down rotation, pegs in a large box with fixed rotation.
\item \textbf{D2 :} Nuts and pegs initialized in a larger box with random top-down rotations.
\end{itemize}

\textbf{Coffee Prep Task}
\textbf{Goal:} Load a mug, open machine, retrieve pod from drawer, and insert pod into machine.

\begin{itemize}
\item \textbf{D0 :} Mug in modest region with fixed rotation, pod in drawer, machine and drawer fixed.
\item \textbf{D1 :} Mug in larger region with random rotation, machine in modest region with top-down rotation variation.
\item \textbf{D2 :} Drawer on right table side, mug on left table side, similar to D0 otherwise.
\end{itemize}

\textbf{Hammer Cleanup Task}
\textbf{Goal:} Open drawer, pick hammer, place into drawer, and close drawer.

\begin{itemize}
\item \textbf{D0 :} Fixed drawer, hammer initialized in a small region with 11 degrees of top-down rotation variation.
\item \textbf{D1 :} Drawer and hammer initialized in large regions, drawer with 60 degrees of top-down rotation variation, hammer with random top-down rotation.
\end{itemize}

\textbf{Mug Cleanup Task}
\textbf{Goal:} Similar to Hammer Cleanup but with a mug and with additional variants.

\begin{itemize}
\item \textbf{D0 :} Fixed drawer, mug initialized in a region with random top-down rotation.
\item \textbf{D1 :} Mug in a region with 60 degrees of top-down rotation variation, drawer also moves in a region with 60 degrees of top-down rotation variation.

\end{itemize}

\section{Derivation of Subtask Segment Transform}
\label{Derivation}
\subsection*{1. Theoretical Goal}
Transform the end-effector trajectory $T_W^E$ from the source demonstration to the current scene's $T_W^E$, satisfying:
\begin{itemize}
    \item Maintain the constancy of $T_O^G$
    \item Adapt to the changes in the current target object pose $T_W^O$
\end{itemize}

\subsection*{2. Key Derivation Steps}

\subsubsection*{(1) Define the Relationship Between Source Demonstration and Current Scene}
In the source demonstration, the relative pose of the gripper to the target object:
\begin{equation}
    T_O^G = (T_{W'}^{O'})^{-1} T_{W'}^{G'}
\end{equation}
In the current scene, this relative pose should remain consistent:
\begin{equation}
    T_O^G = T_{O'}^{G'}
\end{equation}

\subsubsection*{(2) Establish Pose Transformation Chain}
To transform the absolute pose of the source end-effector $T_{W'}^{E'}$ to the current scene, the following relationships are needed:
\begin{align}
    T_W^E &= T_G^E \cdot T_W^G \quad \text{(Current end-effector pose)} \\
    T_{W'}^{E'} &= T_{G'}^{E'} \cdot T_{W'}^{G'} \quad \text{(Source end-effector pose)}
\end{align}
Where $T_G^E$ is not required to be constant (original assumption error), allowing relative motion between the end-effector and the grasped object.

\subsubsection*{(3) Eliminate Intermediate Terms by Combining Equations}
From $T_O^G = T_{O'}^{G'}$, we get:
\begin{equation}
    (T_W^O)^{-1} T_W^G = (T_{W'}^{O'})^{-1} T_{W'}^{G'}
\end{equation}
Rearranging, we get:
\begin{equation}
    T_W^G = T_W^O \cdot (T_{W'}^{O'})^{-1} \cdot T_{W'}^{G'} \quad \text{(Core equation)}
\end{equation}
Substituting $T_W^G$ into the current end-effector pose formula:
\begin{equation}
    T_W^E = T_G^E \cdot \underbrace{T_W^O \cdot (T_{W'}^{O'})^{-1} \cdot T_{W'}^{G'}}_{T_W^G}
\end{equation}
\onecolumn 
\section{Querying Vision-Language Model}
\lstset{ 
    breaklines=true,
    basicstyle={\ttfamily\footnotesize},
    numbers=left, 
    frame=single,
    captionpos=b, 
}

\label{query}
We used Gemini to analyze demonstration videos and to propose constraints for robot tasks. Below are the prompts we used.

\noindent\textbf{Video Analysis Prompt:}
\begin{lstlisting}
"""
Video Analysis:
            In this video, you will see a robotic arm performing a task. The robotic arm has a limited set of predefined skills: it can grasp objects and move them to predefined target locations.

            Please analyze the video and identify the time segments where the robot's actions are too complex for its predefined skills.
            Specifically, please indicate the start and end times (in int seconds) of the video segments where the robot would need to rely on expert demonstration or an expert's data for task completion.

            The video can be seen as a sequence of basic motions, and your task is to identify time segments beyond "grasp" and "move to location".
            Respond with a list of time intervals that require expert knowledge in the format:
            ```json
            [
            {"start": <start_time_in_seconds>, "end": <end_time_in_seconds>},
            {"start": <start_time_in_seconds>, "end": <end_time_in_seconds>},
            ...
            ]
            ```
            For example,
            ```json
            [
            {"start": 2, "end": 4},
            {"start": 7, "end": 11}
            ]
            ```
            """
\end{lstlisting}

\noindent\textbf{Constraint Extraction Prompt:}
\begin{lstlisting}
"""
Constraint Extraction:
## Instructions
Suppose you are controlling a robot within the **RoboSuite simulation environment** to perform a series of object manipulation tasks. Your control is exerted by defining constraint functions in Python.  The task at hand is presented to you via an image depicting the scene, with crucial points (keypoints) highlighted and numbered. Accompanying this is a textual instruction outlining the objective.  For each task, adhere to these steps:

-  Dissect the task into distinct stages. *Crucially, grasping an object must always constitute its own separate stage.* Examples:
    -  "Pour tea from a teapot":
        -  3 stages: "Grasp the teapot", "Align the teapot spout with the cup's opening", and "Pour the liquid".
    - "Place the red block atop the blue block":
        - 3 stages: "Grasp the red block", "Lift and position the red block above the blue block", and "Release the red block onto the blue block."
    -  "Reorient a bouquet and place it upright in a vase":
        -  3 stages: "Grasp the bouquet's stem", "Reorient the bouquet to an upright position", and "Maintain upright orientation while lowering it into the vase."
- For every stage, formulate two types of constraints: "sub-goal constraints" and "path constraints."  "Sub-goal constraints" define conditions that *must be met at the conclusion of the stage*. "Path constraints" dictate conditions that *must hold true throughout the stage's execution*. Examples:
  - "Pour tea from a teapot":
    - "Grasp teapot" stage:
      - 1 sub-goal constraint: "The robot's end-effector must be aligned with the teapot's handle."
      - 0 path constraints.
    - "Align teapot with cup opening" stage:
      - 1 sub-goal constraint: "The teapot's spout should be positioned 10cm above the cup's opening."
      - 2 path constraints: "The robot must maintain its grasp on the teapot handle", "The teapot must remain upright to prevent spillage."
    - "Pour liquid" stage:
      - 2 sub-goal constraints: "The teapot's spout should be 5cm above the cup's opening", "The teapot must be tilted to initiate pouring."
      - 2 path constraints: "The robot must maintain its grasp on the teapot handle", "The teapot's spout must be directly above the cup's opening."
  - "Place the red block atop the blue block":
    -  "Grasp red block" stage:
      -  1 sub-goal constraint: "Align the end-effector with the red block."
      -  0 path constraints
    - "Lift and position/Release":
      - 1 sub-goal constraint: "the red block is 10cm on top of the blue block"
      - 1 path constraint: "The robot must still be grasping the red block".
  - "Reorient a bouquet and place it upright in a vase":
    -  "Grasp bouquet" stage:
       - 1 sub-goal constraint: "Align the end-effector with the bouquet stem."
       - 0 path constraints.
    - "Reorient bouquet" stage:
      - 1 sub-goal constraint: "The bouquet is oriented vertically (parallel to the z-axis)."
      - 1 path constraint: "The robot must maintain its grasp on the bouquet stem."
    -  "Maintain upright/Lower into vase" stage:
      -  2 sub-goal constraints: "The bouquet must remain upright (parallel to the z-axis)", "The base of the bouquet is 20cm above the vase opening."
      -  1 path constraint: "The robot must maintain its grasp on the bouquet stem."
-  Identify and record the keypoints that need to be grasped during any grasping stages using the `grasp_keypoints` variable.
-  Specify the stage at the *end* of which the robot should release the grasped keypoints by setting the `release_keypoints` variable.

**Important Notes:**

-  Each constraint function accepts a dummy end-effector position (a 3D point) and an array of keypoint positions as input.  It returns a numerical cost; the constraint is considered satisfied if this cost is less than or equal to zero.
-  For each stage, you can define zero or more sub-goal constraints and zero or more path constraints.
-  *Strictly avoid using "if" statements within your constraint functions.*
- *Refrain from using path constraints when dealing with deformable objects like clothing or towels.*
-  *You do not need to implement collision avoidance. Concentrate solely on the actions required to successfully complete the task.*
-  Input parameters for the constraint functions are:
  -  `end_effector`: A NumPy array of shape `(3,)` representing the 3D position of the robot's end-effector.
  -  `keypoints`: A NumPy array of shape `(K, 3)` representing the 3D positions of the keypoints.
-  For any path constraint that necessitates the robot maintaining its grasp on a keypoint `i`, you can utilize the provided helper function `get_grasping_cost_by_keypoint_idx(i)`. Simply `return get_grasping_cost_by_keypoint_idx(i)` where `i` is the index of the relevant keypoint.
-  Within each function, you are permitted to use: native Python functions, any NumPy functions, and the `get_grasping_cost_by_keypoint_idx` function.
-  For a grasping stage, you should *only* define *one* sub-goal constraint. This constraint should associate the end-effector with the appropriate keypoint. No path constraints are needed for grasping stages.
-  To move a keypoint, the object associated with that keypoint *must* have been grasped in a preceding stage.
- The robot is limited to grasping *one* object at a time.
-  *Grasping must always be a distinct stage, separate from other actions.*
-  You can use two keypoints to define a vector. This vector can then be used to specify a rotation (by calculating the angle between the vector and a fixed axis, for example).
- Multiple keypoints can be used to define a surface or a volume.
-  The keypoint indices marked on the image, and in the `keypoints` array, start at 0.
- To specify a point `i` relative to another point `j`, your function should define an `offsetted_point` variable. This variable is calculated by adding the desired delta (offset) to keypoint `j`. Then, calculate the norm of the difference between keypoint `i` and this `offsetted_point`.
- If you need to specify a location that is *not* explicitly marked by a keypoint, try to define it using multiple keypoints (e.g., by taking the mean of several keypoints if the desired location lies at their center).

**Structure your output in a single python code block as follows:**
```python

# Explanation of the stages involved in the task.
# ...

num_stages = ?

### stage 1 sub-goal constraints (if any)
def stage1_subgoal_constraint1(end_effector, keypoints):
    """Explanation of the constraint."""
    ...
    return cost
# Add more sub-goal constraints if needed
...

### stage 1 path constraints (if any)
def stage1_path_constraint1(end_effector, keypoints):
    """Explanation of the constraint."""
    ...
    return cost
# Add more path constraints if needed
...

# repeat for more stages
...

"""
Summarize keypoints to be grasped in all grasping stages.
The length of the list should be equal to the number of stages.
For grapsing stage, write the keypoint index. For non-grasping stage, write -1.
"""
grasp_keypoints = [?, ..., ?]

"""
Summarize at **the end of which stage** the robot should release the keypoints.
The keypoint indices must appear in an earlier stage as defined in `grasp_keypoints` (i.e., a keypoint can only be released only if it has been grasped previously).
Only release object when it's necessary to complete the task, e.g., drop bouquet in the vase.
The length of the list should be equal to the number of stages.
If a keypoint is to be released at the end of a stage, write the keypoint index at the corresponding location. Otherwise, write -1.
"""
release_keypoints = [?, ..., ?]

```

## Query
Query Task: "pick up the red cube and lift it up"
Query Image:
Query Video:
}
"""
\end{lstlisting}

%% file: main.bbl
\begin{thebibliography}{34}
\providecommand{\natexlab}[1]{#1}
\providecommand{\url}[1]{\texttt{#1}}
\expandafter\ifx\csname urlstyle\endcsname\relax
  \providecommand{\doi}[1]{doi: #1}\else
  \providecommand{\doi}{doi: \begingroup \urlstyle{rm}\Url}\fi

\bibitem[Achiam et~al.(2023)Achiam, Adler, Agarwal, Ahmad, Akkaya, Aleman, Almeida, Altenschmidt, Altman, Anadkat, et~al.]{openaiGPT4TechnicalReport2024a}
Josh Achiam, Steven Adler, Sandhini Agarwal, Lama Ahmad, Ilge Akkaya, Florencia~Leoni Aleman, Diogo Almeida, Janko Altenschmidt, Sam Altman, Shyamal Anadkat, et~al.
\newblock Gpt-4 technical report.
\newblock \emph{arXiv preprint arXiv:2303.08774}, 2023.

\bibitem[Chi et~al.()Chi, Xu, Feng, Cousineau, Du, Burchfiel, Tedrake, and Song]{chiDiffusionPolicyVisuomotor2024}
Cheng Chi, Zhenjia Xu, Siyuan Feng, Eric Cousineau, Yilun Du, Benjamin Burchfiel, Russ Tedrake, and Shuran Song.
\newblock Diffusion policy: Visuomotor policy learning via action diffusion.
\newblock \emph{The International Journal of Robotics Research}, page 02783649241273668.

\bibitem[Di~Palo and Johns(2022)]{paloLearningMultiStageTasks2021}
Norman Di~Palo and Edward Johns.
\newblock Learning multi-stage tasks with one demonstration via self-replay.
\newblock In \emph{Conference on Robot Learning}, pages 1180--1189. PMLR, 2022.

\bibitem[Diankov(2010)]{diankov2010automated}
Rosen Diankov.
\newblock \emph{Automated construction of robotic manipulation programs}.
\newblock PhD thesis, Carnegie Mellon University, USA, 2010.

\bibitem[Dosovitskiy et~al.(2020)Dosovitskiy, Beyer, Kolesnikov, Weissenborn, Zhai, Unterthiner, Dehghani, Minderer, Heigold, Gelly, et~al.]{dosovitskiy2020image}
Alexey Dosovitskiy, Lucas Beyer, Alexander Kolesnikov, Dirk Weissenborn, Xiaohua Zhai, Thomas Unterthiner, Mostafa Dehghani, Matthias Minderer, G Heigold, S Gelly, et~al.
\newblock An image is worth 16x16 words: Transformers for image recognition at scale.
\newblock In \emph{International Conference on Learning Representations}, 2020.

\bibitem[Garrett et~al.(2024)Garrett, Mandlekar, Wen, and Fox]{garrettSkillMimicGenAutomatedDemonstration2024}
Caelan Garrett, Ajay Mandlekar, Bowen Wen, and Dieter Fox.
\newblock Skillmimicgen: Automated demonstration generation for efficient skill learning and deployment.
\newblock \emph{arXiv preprint arXiv:2410.18907}, 2024.

\bibitem[Ha et~al.(2023)Ha, Florence, and Song]{haScalingDistillingLanguageGuided2023}
Huy Ha, Pete Florence, and Shuran Song.
\newblock Scaling up and distilling down: Language-guided robot skill acquisition.
\newblock In \emph{Conference on Robot Learning}, pages 3766--3777. PMLR, 2023.

\bibitem[Hu et~al.(2023)Hu, Lin, Zhang, Yi, and Gao]{huLookYouLeap2023}
Yingdong Hu, Fanqi Lin, Tong Zhang, Li Yi, and Yang Gao.
\newblock Look before you leap: Unveiling the power of gpt-4v in robotic vision-language planning.
\newblock \emph{arXiv preprint arXiv:2311.17842}, 2023.

\bibitem[Huang et~al.(2024)Huang, Lin, Hu, Wang, and Gao]{huangCoPaGeneralRobotic}
Haoxu Huang, Fanqi Lin, Yingdong Hu, Shengjie Wang, and Yang Gao.
\newblock Copa: General robotic manipulation through spatial constraints of parts with foundation models.
\newblock In \emph{2024 IEEE/RSJ International Conference on Intelligent Robots and Systems (IROS)}, pages 9488--9495. IEEE, 2024.

\bibitem[Huang et~al.(2023{\natexlab{a}})Huang, Jiang, Dong, Qiao, Gao, and Li]{huangInstruct2ActMappingMultimodality2023}
Siyuan Huang, Zhengkai Jiang, Hao Dong, Yu Qiao, Peng Gao, and Hongsheng Li.
\newblock Instruct2act: Mapping multi-modality instructions to robotic actions with large language model.
\newblock \emph{arXiv preprint arXiv:2305.11176}, 2023{\natexlab{a}}.

\bibitem[Huang et~al.()Huang, Wang, Li, Zhang, and Fei-Fei]{huangReKepSpatioTemporalReasoning}
Wenlong Huang, Chen Wang, Yunzhu Li, Ruohan Zhang, and Li Fei-Fei.
\newblock Rekep: Spatio-temporal reasoning of relational keypoint constraints for robotic manipulation.
\newblock In \emph{8th Annual Conference on Robot Learning}.

\bibitem[Huang et~al.(2023{\natexlab{b}})Huang, Wang, Zhang, Li, Wu, and Fei-Fei]{huangVoxPoserComposable3D2023}
Wenlong Huang, Chen Wang, Ruohan Zhang, Yunzhu Li, Jiajun Wu, and Li Fei-Fei.
\newblock Voxposer: Composable 3d value maps for robotic manipulation with language models.
\newblock In \emph{Conference on Robot Learning}, pages 540--562. PMLR, 2023{\natexlab{b}}.

\bibitem[Jiang et~al.(2022)Jiang, Gupta, Zhang, Wang, Dou, Chen, Fei-Fei, Anandkumar, Zhu, and Fan]{jiangVIMAGeneralRobot2023}
Yunfan Jiang, Agrim Gupta, Zichen Zhang, Guanzhi Wang, Yongqiang Dou, Yanjun Chen, Li Fei-Fei, Anima Anandkumar, Yuke Zhu, and Linxi Fan.
\newblock Vima: General robot manipulation with multimodal prompts.
\newblock \emph{arXiv preprint arXiv:2210.03094}, 2\penalty0 (3):\penalty0 6, 2022.

\bibitem[Jiang et~al.()Jiang, Xie, Lin, Xu, Wan, Mandlekar, Fan, and Zhu]{jiangDexMimicGenAutomatedData2024}
Zhenyu Jiang, Yuqi Xie, Kevin Lin, Zhenjia Xu, Weikang Wan, Ajay Mandlekar, Linxi Fan, and Yuke Zhu.
\newblock Dexmimicgen: Automated data generation for bimanual dexterous manipulation via imitation learning.
\newblock In \emph{CoRL Workshop on Learning Robot Fine and Dexterous Manipulation: Perception and Control}.

\bibitem[Jin et~al.(2024)Jin, Li, Yong, Shi, Hao, Sun, Zhang, and Fang]{jinRobotGPTRobotManipulation2023}
Yixiang Jin, Dingzhe Li, A Yong, Jun Shi, Peng Hao, Fuchun Sun, Jianwei Zhang, and Bin Fang.
\newblock Robotgpt: Robot manipulation learning from chatgpt.
\newblock \emph{IEEE Robotics and Automation Letters}, 9\penalty0 (3):\penalty0 2543--2550, 2024.

\bibitem[Johns(2021)]{johns2021coarse}
Edward Johns.
\newblock Coarse-to-fine imitation learning: Robot manipulation from a single demonstration.
\newblock In \emph{2021 IEEE international conference on robotics and automation (ICRA)}, pages 4613--4619. IEEE, 2021.

\bibitem[Kim et~al.(2023)Kim, Kim, Kim, Min, and Choi]{kimContextAwarePlanningEnvironmentAware2024}
Byeonghwi Kim, Jinyeon Kim, Yuyeong Kim, Cheolhong Min, and Jonghyun Choi.
\newblock Context-aware planning and environment-aware memory for instruction following embodied agents.
\newblock In \emph{Proceedings of the IEEE/CVF International Conference on Computer Vision}, pages 10936--10946, 2023.

\bibitem[Liang et~al.(2023)Liang, Huang, Xia, Xu, Hausman, Ichter, Florence, and Zeng]{liangCodePoliciesLanguage2023}
Jacky Liang, Wenlong Huang, Fei Xia, Peng Xu, Karol Hausman, Brian Ichter, Pete Florence, and Andy Zeng.
\newblock Code as policies: Language model programs for embodied control.
\newblock In \emph{2023 IEEE International Conference on Robotics and Automation (ICRA)}, pages 9493--9500. IEEE, 2023.

\bibitem[Lin et~al.(2024)Lin, Hu, Sheng, Wen, You, and Gao]{linDATASCALINGLAWS}
Fanqi Lin, Yingdong Hu, Pingyue Sheng, Chuan Wen, Jiacheng You, and Yang Gao.
\newblock Data scaling laws in imitation learning for robotic manipulation.
\newblock \emph{arXiv preprint arXiv:2410.18647}, 2024.

\bibitem[Mandlekar et~al.(2022)Mandlekar, Xu, Wong, Nasiriany, Wang, Kulkarni, Fei-Fei, Savarese, Zhu, and Mart{\'\i}n-Mart{\'\i}n]{mandlekar2022matters}
Ajay Mandlekar, Danfei Xu, Josiah Wong, Soroush Nasiriany, Chen Wang, Rohun Kulkarni, Li Fei-Fei, Silvio Savarese, Yuke Zhu, and Roberto Mart{\'\i}n-Mart{\'\i}n.
\newblock What matters in learning from offline human demonstrations for robot manipulation.
\newblock In \emph{Conference on Robot Learning}, pages 1678--1690. PMLR, 2022.

\bibitem[Mandlekar et~al.(2023{\natexlab{a}})Mandlekar, Garrett, Xu, and Fox]{mandlekarHumanintheLoopTaskMotion2023}
Ajay Mandlekar, Caelan~Reed Garrett, Danfei Xu, and Dieter Fox.
\newblock Human-in-the-loop task and motion planning for imitation learning.
\newblock In \emph{Conference on Robot Learning}, pages 3030--3060. PMLR, 2023{\natexlab{a}}.

\bibitem[Mandlekar et~al.(2023{\natexlab{b}})Mandlekar, Nasiriany, Wen, Akinola, Narang, Fan, Zhu, and Fox]{mandlekarMimicGenDataGeneration2023}
Ajay Mandlekar, Soroush Nasiriany, Bowen Wen, Iretiayo Akinola, Yashraj Narang, Linxi Fan, Yuke Zhu, and Dieter Fox.
\newblock Mimicgen: A data generation system for scalable robot learning using human demonstrations.
\newblock In \emph{Conference on Robot Learning}, pages 1820--1864. PMLR, 2023{\natexlab{b}}.

\bibitem[Mu et~al.(2024)Mu, Chen, Peng, Chen, Gao, Zou, Lin, Xie, and Luo]{muRoboTwinDualArmRobot2024}
Yao Mu, Tianxing Chen, Shijia Peng, Zanxin Chen, Zeyu Gao, Yude Zou, Lunkai Lin, Zhiqiang Xie, and Ping Luo.
\newblock Robotwin: Dual-arm robot benchmark with generative digital twins (early version).
\newblock \emph{arXiv preprint arXiv:2409.02920}, 2024.

\bibitem[Pomerleau(1988)]{pomerleau1988alvinn}
Dean~A Pomerleau.
\newblock Alvinn: An autonomous land vehicle in a neural network.
\newblock \emph{Advances in neural information processing systems}, 1, 1988.

\bibitem[Tang et~al.(2024)Tang, Rajkumar, Zhou, Walke, Levine, and Fang]{tangKALIEFineTuningVisionLanguage2024}
Grace Tang, Swetha Rajkumar, Yifei Zhou, Homer~Rich Walke, Sergey Levine, and Kuan Fang.
\newblock Kalie: Fine-tuning vision-language models for open-world manipulation without robot data.
\newblock \emph{arXiv preprint arXiv:2409.14066}, 2024.

\bibitem[Team et~al.(2023)Team, Anil, Borgeaud, Alayrac, Yu, Soricut, Schalkwyk, Dai, Hauth, Millican, et~al.]{geminiteamGeminiFamilyHighly2024}
Gemini Team, Rohan Anil, Sebastian Borgeaud, Jean-Baptiste Alayrac, Jiahui Yu, Radu Soricut, Johan Schalkwyk, Andrew~M Dai, Anja Hauth, Katie Millican, et~al.
\newblock Gemini: a family of highly capable multimodal models.
\newblock \emph{arXiv preprint arXiv:2312.11805}, 2023.

\bibitem[Team et~al.(2024)Team, Georgiev, Lei, Burnell, Bai, Gulati, Tanzer, Vincent, Pan, Wang, et~al.]{teamGemini15Unlocking2024}
Gemini Team, Petko Georgiev, Ving~Ian Lei, Ryan Burnell, Libin Bai, Anmol Gulati, Garrett Tanzer, Damien Vincent, Zhufeng Pan, Shibo Wang, et~al.
\newblock Gemini 1.5: Unlocking multimodal understanding across millions of tokens of context.
\newblock \emph{arXiv preprint arXiv:2403.05530}, 2024.

\bibitem[Vosylius and Johns(2023)]{vosylius2023start}
Vitalis Vosylius and Edward Johns.
\newblock Where to start? transferring simple skills to complex environments.
\newblock In \emph{Conference on Robot Learning}, pages 471--481. PMLR, 2023.

\bibitem[Wang et~al.(2024{\natexlab{a}})Wang, Zhang, Dong, Fang, and Feng]{wangVLMSeeRobot2024}
Beichen Wang, Juexiao Zhang, Shuwen Dong, Irving Fang, and Chen Feng.
\newblock Vlm see, robot do: Human demo video to robot action plan via vision language model.
\newblock \emph{arXiv preprint arXiv:2410.08792}, 2024{\natexlab{a}}.

\bibitem[Wang et~al.(2024{\natexlab{b}})Wang, Xian, Chen, Wang, Wang, Fragkiadaki, Erickson, Held, and Gan]{wangRoboGenUnleashingInfinite2023}
Yufei Wang, Zhou Xian, Feng Chen, Tsun-Hsuan Wang, Yian Wang, Katerina Fragkiadaki, Zackory Erickson, David Held, and Chuang Gan.
\newblock Robogen: towards unleashing infinite data for automated robot learning via generative simulation.
\newblock In \emph{Proceedings of the 41st International Conference on Machine Learning}, pages 51936--51983, 2024{\natexlab{b}}.

\bibitem[Wu et~al.(2023)Wu, Antonova, Kan, Lepert, Zeng, Song, Bohg, Rusinkiewicz, and Funkhouser]{wuTidyBotPersonalizedRobot}
Jimmy Wu, Rika Antonova, Adam Kan, Marion Lepert, Andy Zeng, Shuran Song, Jeannette Bohg, Szymon Rusinkiewicz, and Thomas Funkhouser.
\newblock Tidybot: Personalized robot assistance with large language models.
\newblock \emph{Autonomous Robots}, 47\penalty0 (8):\penalty0 1087--1102, 2023.

\bibitem[Zhang et~al.()Zhang, Hu, You, and Gao]{zhangLeveragingLocalityBoost2024}
Tong Zhang, Yingdong Hu, Jiacheng You, and Yang Gao.
\newblock Leveraging locality to boost sample efficiency in robotic manipulation.
\newblock In \emph{8th Annual Conference on Robot Learning}.

\bibitem[Zhao et~al.()Zhao, Tompson, Driess, Florence, Ghasemipour, Finn, and Wahid]{zhaoALOHAUnleashedSimple}
Tony~Z Zhao, Jonathan Tompson, Danny Driess, Pete Florence, Seyed Kamyar~Seyed Ghasemipour, Chelsea Finn, and Ayzaan Wahid.
\newblock Aloha unleashed: A simple recipe for robot dexterity.
\newblock In \emph{8th Annual Conference on Robot Learning}.

\bibitem[Zhou et~al.(2024)Zhou, Su, Chi, Zhang, Wang, Huang, Sheng, and Wang]{zhouCodeasMonitorConstraintawareVisual2024}
Enshen Zhou, Qi Su, Cheng Chi, Zhizheng Zhang, Zhongyuan Wang, Tiejun Huang, Lu Sheng, and He Wang.
\newblock Code-as-monitor: Constraint-aware visual programming for reactive and proactive robotic failure detection.
\newblock \emph{arXiv preprint arXiv:2412.04455}, 2024.

\end{thebibliography}
